\documentclass[journal]{IEEEtran}
\usepackage{graphicx}

\usepackage{cite}

\ifCLASSINFOpdf
\else
\fi
\usepackage{amsmath}
\usepackage{framed,multirow}

\usepackage{algorithm}
\usepackage{algorithmic}
\begin{document}
%
\title{Panoptic Feature Fusion Net: A Novel Instance Segmentation Paradigm for Biomedical and Biological Images}
%
%
%

\author{Dongnan~Liu, 
        Donghao~Zhang, 
        Yang~Song,~\IEEEmembership{Member,~IEEE,}
        Heng~Huang,
        and~Weidong~Cai,~\IEEEmembership{Member,~IEEE}
\thanks{D. Liu, D. Zhang, and W. Cai are with the School of Computer Science, University of Sydney, NSW,
2008 Australia (e-mail: dliu5812@uni.sydney.edu.au).}
\thanks{Y. Song is with the School of Computer Science and Engineering, University of New South Wales, NSW, 2052 Australia.}
\thanks{H. Huang is with the Department of Electrical and Computer Engineering, University of Pittsburgh, PA, 15261 USA}}

\maketitle

\begin{abstract}

Instance segmentation is an important task for biomedical and biological image analysis. Due to the complicated background components, the high variability of object appearances, numerous overlapping objects, and ambiguous object boundaries, this task still remains challenging. Recently, deep learning based methods have been widely employed to solve these problems and can be categorized into proposal-free and proposal-based methods. However, both proposal-free and proposal-based methods suffer from information loss, as they focus on either global-level semantic or local-level instance features. To tackle this issue, we present a Panoptic Feature Fusion Net (PFFNet) that unifies the semantic and instance features in this work. Specifically, our proposed PFFNet contains a residual attention feature fusion mechanism to incorporate the instance prediction with the semantic features, in order to facilitate the semantic contextual information learning in the instance branch. Then, a mask quality sub-branch is designed to align the confidence score of each object with the quality of the mask prediction. Furthermore, a consistency regularization mechanism is designed between the semantic segmentation tasks in the semantic and instance branches, for the robust learning of both tasks. Extensive experiments demonstrate the effectiveness of our proposed PFFNet, which outperforms several state-of-the-art methods on various biomedical and biological datasets.

\end{abstract}


%
\IEEEpeerreviewmaketitle

\section{Introduction}

\IEEEPARstart{I}{nstance} segmentation is a prerequisite step for biomedical and biological image processing, which not only assigns a class label for each pixel but also separates each object within the same class. By assigning a unique ID for every single object, the morphology, spatial locations, and distribution of the objects can be further studied to analyze the biological behaviors from the given images. In the digital pathology domain, the nuclear pleomorphism (size and shape) contributes to the tumor and cancer grading, and the spatial arrangement of cancer nuclei facilitates the understanding of cancer prognostic predictions \cite{elston1991pathological,clayton1991pathologic,basavanhally2011multi,xing2016robust}. In the plant and agriculture study, analyzing each distinguished leaf in plant images enables the experts to learn about the plant phenotype including the number of leaves, maturity condition, and its similar cultivars, which serves as the key factor of understanding plant function and growth condition \cite{minervini2016finely,scharr2016leaf,scharr2014annotated}. Traditional manual assessment for biomedical and biological image instance segmentation is not suitable for current practice, as it is labor-intensive and time-consuming. Additionally, limitations of objective and reproducibility are unavoidable due to the intra- and inter-observer variability \cite{llewellyn2000observer}. To this end, automatic and accurate methods for instance segmentation in biology images are necessary and in high demand.


There still remain some challenges in instance segmentation tasks for biomedical and biological images. First, some background structures have a similar appearance to the foreground object, such as cytoplasm or stroma in histopathology images. Therefore, methods relying on thresholding are ineffective. Second, within the same dataset, the objects in different images have large variability in size, shape, texture, and intensity. It is caused by the various biological structures and activities when acquiring different images \cite{song2018contour,payer2019segmenting}. Third, there are clusters of objects overlapping with each other. The boundaries between these touching objects are ambiguous due to nonuniform staining absorption and similar object intensity. This might result in segmenting several objects into a single one. In order to tackle these issues, deep learning based methods are prevalent and effective by learning from feature representations.



CNN based instance segmentation methods can be categorized into two types: proposal-free and proposal-based methods. For the proposal-free instance segmentation methods, each pixel is firstly assigned a class label with a semantic segmentation model. The post-processing steps are then employed to separate each foreground object within the same category, according to their morphology characteristic, structures, and spatial arrangement \cite{kumar2017dataset,naylor2018segmentation,payer2018instance,de2017semantic,chen2017dcan}. Although post-processing among these methods is capable of separating the connected components, they still suffer from artificial boundaries during overlapping object segmentation. Even though \cite{kumar2017dataset,chen2017dcan,zhang2018nuclei} focus on boundaries learning at the semantic segmentation stage, the global contextual information is still not enough to separate the touching objects, especially when their borders become unclear. On the other hand, the proposal-based instance segmentation methods incorporate the detection task with the segmentation task \cite{he2017mask,liu2018path}. First, the spatial location for each object is detected as a bounding box. Then, a mask generator is further employed to segment each object within the corresponding predicted bounding box. By detecting and segmenting every single object separately, the proposal-based methods are capable of separating the touching objects. However, they are limited as there is a lack of global semantic information between the foreground and background.

For the instance segmentation tasks, both the global semantic and local instance information is important. The global semantic information indicates the useful clues in the scene context, such as the relationship between the foreground and background and the spatial distribution of all the foreground objects. On the other hand, local-level instance information describes the spatial location and detailed contour for every single object. To integrate the benefits of the global and local features, panoptic segmentation \cite{kirillov2018panoptic}, reconciliation of the semantic and instance segmentation, has been proposed. In~\cite{kirillov2018panoptic}, the predictions from two separately trained semantic and instance segmentation branches are fused together to analyze the panoptic level segmentation. Without sharing components between the two branches, training~\cite{kirillov2018panoptic} incurs a large computational cost~\cite{kirillov2019panoptic}. In addition, the analysis in~\cite{kirillov2019panoptic} indicates that jointly training a network for the two tasks achieves better performance than training them independently. To this end, Panoptic PFN~\cite{kirillov2019panoptic} is proposed to jointly train the semantic and instance segmentation branches by sharing the same ResNet backbone, which has achieved state-of-the-art performance on panoptic segmentation as well as maintained memory efficiency.

Based on \cite{kirillov2018panoptic} and \cite{kirillov2019panoptic}, we previously proposed \cite{zhang2018panoptic} and \cite{liu2019nuclei} for nuclei instance segmentation in histopathology images. Motivated by jointly analyzing the semantic and instance segmentation tasks in~\cite{kirillov2018panoptic}, we designed the Cell R-CNN \cite{zhang2018panoptic} to simultaneously process the global and local information in the histopathology images. Different from the two separately optimized branches in \cite{kirillov2018panoptic}, our Cell R-CNN~\cite{zhang2018panoptic} proposed to jointly train the two branches with a shared backbone model. In order to further facilitate the semantic-level contextual learning in the instance segmentation model, our IJCAI work \cite{liu2019nuclei} was proposed to induce the instance branch to learn directly about the semantic-level features. As the extension of Cell R-CNN, we refer to \cite{liu2019nuclei} as Cell R-CNN V2 in the following sections. In Cell R-CNN V2, we firstly introduce a new semantic segmentation prediction from the instance branch. Then a feature fusion mechanism to incorporate the feature maps is designed to induce the semantic feature learning in the decoder of the instance segmentation branch, by integrating the mask prediction of the instance branch with that of the semantic branch. In addition, a dual-model mask generator is proposed for instance mask segmentation, in order to prevent information loss. Compared with the Panoptic FPN~\cite{kirillov2019panoptic}, which only jointly optimized the semantic and instance segmentation branches with a shared backbone, our Cell R-CNN V2 directly integrated the features from the two branches, to further induce the semantic feature learning in the instance branch.


In this work, we propose a Panoptic Feature Fusion Net (PFFNet), which further extends our preliminary Cell R-CNN V2 \cite{liu2019nuclei} by addressing several remaining problems. First, the feature fusion mechanism in \cite{liu2019nuclei} directly replaces the part of the feature map in the semantic segmentation branch with those from the output of the mask generator. Although the mask predictions from the instance branch interpret more instance-level features than the semantic branch, the global contextual features from the semantic segmentation prediction are also important. To this end, we propose a residual attention feature fusion mechanism (RAFF) in this work, to replace the previous feature fusion mechanism. In our newly proposed RAFF, the local features from the instance branch are integrated with the global semantic features, without deprecating any semantic-level features. Second, two semantic segmentation tasks with the same ground truth are optimized together in the overall architecture of \cite{liu2019nuclei}. In order to facilitate the robust learning of two segmentation tasks, we add a semantic consistency regularization between them to enforce the two semantic predictions from two different branches as similar as possible. In addition, there remain some low-quality mask predictions with an unexpected high classification score in the traditional Mask R-CNN, as mentioned in \cite{huang2019mask}. It would be harmful to the segmentation accuracy if treating these poorly generated results as the ones with high confidence. To this end, we propose a new mask quality sub-branch in this work, by learning an auxiliary quality score of each mask prediction based on the Dice score and Intersection-over-Union (IoU) score. During inference, the classification score of each mask is re-weighted through multiplication by its corresponding mask quality score.

The PFFNet proposed in this manuscript is an extension of Cell R-CNN V2, and can therefore also be named Cell R-CNN V3. In line with our previous Cell R-CNN V2 and Cell R-CNN, we are the first to employ the panoptic segmentation idea on biomedical and biological image analysis, to the best of our knowledge. Overall, the contributions of this work compared with Cell R-CNN V2 are summarized as follows:
\begin{itemize}
\item We design a residual attention feature fusion mechanism to integrate the features of each detected object in the semantic and instance levels. 
\item We design a semantic task consistency mechanism to regularize the semantic segmentation tasks training for robustness.
\item We design an extra mask quality sub-branch to ensure the mask segmentation quality for each object is compatible with its confidence score.
\item Our proposed Panoptic Feature Fusion Net is validated on the instance segmentation tasks for various biomedical and biological datasets, including histopathology images, fluorescence microscopy images, and plant phenotyping images. Our results for all metrics outperform the state-of-the-art methods by a large margin.
\end{itemize}


\section{Related work}

Instance segmentation for biomedical and biological images is widely studied, ranging from the handcrafted feature-based methods to the learning-based methods. In order to emphasize the contributions of our proposed PFFNet, we mainly focus on the literature of deep learning based instance segmentation methods, which can be grouped into two classes: the proposal-free and proposal-based methods. 

\subsection{Proposal-free Instance Segmentation}

Proposal-free instance segmentation methods are mainly based on the morphology and spatial relationship of all the objects in the images. For example, object boundary is an important feature for separating the touching object. In \cite{kumar2017dataset,chen2017dcan,zhang2018nuclei,bai2017deep}, the instances are separated according to the probability map for the foreground objects and their boundaries. Similarly, \cite{kulikov2018instance} separates each instance according to the distance between the two connected components. Additionally, post-processing methods are employed to separate the touching objects based on the semantic segmentation predictions, such as conditional region growing algorithm \cite{kumar2017dataset}, morphological dynamics algorithms \cite{kumar2017dataset}, and watershed algorithm \cite{zhang2018nuclei,chen2017dcan}. In addition to the traditional classification-based segmentation methods, regression-based methods are also widely employed. In \cite{naylor2018segmentation}, a distance transform map describing the distance between each pixel and its nearest background pixel is predicted, with a regression CNN architecture. To obtain the instance segmentation map directly, \cite{payer2018instance,kulikov2020instance,de2017semantic} employ the clustering algorithm on the high dimensional embeddings predicted from the deep regression CNN model. Based on adversarial learning architecture, Zhang et al \cite{zhang2018nuclei} proposed an image-to-image translation method for a more accurate probability map compared with the classification-based method. 

\subsection{Proposal-based Instance Segmentation}

Compared with proposal-free instance segmentation methods, the proposal-based methods predict the mask segmentation for each object based on the predictions of their corresponding locations in the whole image \cite{ren2017end,chen2018masklab}. One fundamental proposal-based instance segmentation method is Mask R-CNN \cite{he2017mask}. Based on the high-dimensional feature maps from the backbone CNN network, Mask R-CNN firstly generates regions of interest (ROIs) containing the foreground objects with a region proposal network (RPN). After aligning the ROIs to the same size, a box sub-branch and a mask sub-branch are employed to predict the coordinate, class label, and mask prediction for each ROI. With the help of the local-level information from the spatial locations of the instances, Mask R-CNN achieved state-of-the-art performance compared with the traditional box-free methods. Following the Mask R-CNN, other methods were further proposed with a higher accuracy: \cite{liu2018path} proposed a path aggregate backbone to preserve the feature maps at high resolutions, \cite{huang2019mask} added a branch for mask IoU score prediction based on the mask prediction on the original Mask R-CNN, and  \cite{cai2019cascade} employed a cascade connection of several bounding box and mask prediction sub-branches.

Although the proposal-based instance segmentation methods achieve higher performance compared with the proposal-free methods by processing each object separately, their effectiveness is still limited due to the lack of the semantic-level global information on the context of the whole images. To tackle this issue, panoptic segmentation was recently proposed to jointly process the foreground things and the background stuff \cite{kirillov2018panoptic}, by incorporating the semantic segmentation with the instance segmentation. Inspired by this joint segmentation idea, \cite{saleh2018effective} fuses the instance segmentation result for foreground objects with the semantic segmentation result for the background for urban scene semantic segmentation. However, the instance branch and semantic branch are trained separately in previous work. In \cite{de2018panoptic,kirillov2019panoptic}, both the instance and semantic segmentation branches are trained together by sharing the same backbone module. Then, the losses of the two branches are summed together for back propagation to optimize the parameters of the whole framework. Later, more methods for fusing the results of things and stuff are proposed. In \cite{li2019attention}, attention mechanism is employed to fuse the proposals and masks from the instance branch with the feature map from the semantic branch. \cite{liu2019end} proposed a spatial ranking module to separate the overlapping objects from different categories by fusing the semantic segmentation predictions with the instance segmentation ones. 


Similar to the jointly learning paradigm in the panoptic segmentation, combining the semantic segmentation task of the proposal-based instance segmentation also enables the model to achieve higher performance by learning the auxiliary semantic-level contextual information. In \cite{chen2019hybrid}, the semantic segmentation prediction is fused with the proposed hybrid cascade instance segmentation architecture to make the architecture manipulate the global semantic features and achieve state-of-the-art performance compared with previous instance segmentation methods. In medical analysis tasks, we previously proposed Cell R-CNN \cite{zhang2018panoptic} to induce the encoder of the instance segmentation to learn semantic-level information by jointly training a semantic segmentation network and a Mask R-CNN with a shared backbone network. With the help of the semantic-level contextual information, Cell R-CNN outperforms Mask R-CNN in the nuclei segmentation tasks on histopathology images. However, the decoder of the Cell R-CNN only learns the semantic features indirectly, which still makes the model lack global information during inference. In Cell R-CNN V2 \cite{liu2019nuclei}, we, therefore, designed a feature fusion module to incorporate the feature maps from the semantic segmentation branch and the instance segmentation branch during the training phase. By retaining semantic-level features in the encoder and decoder of the instance segmentation model, our previous work \cite{liu2019nuclei} achieved state-of-the-art performance on several nuclei instance segmentation tasks under both object- and pixel-level metrics.

\section{Panoptic Feature Fusion Net}

\begin{figure*}[!t]
\centering
\includegraphics[width=0.75\textwidth]{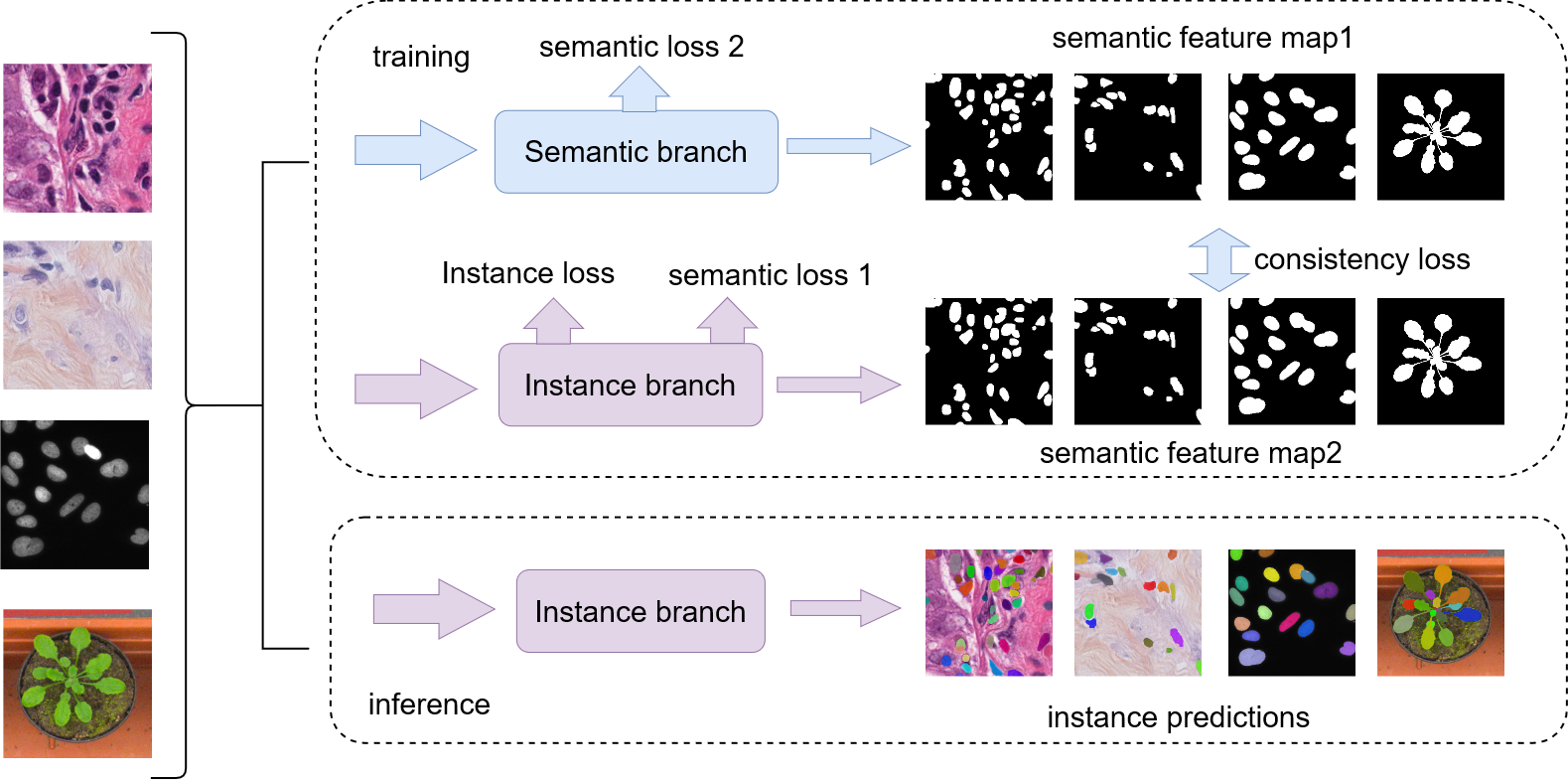}
\caption{Overview of our proposed Panoptic Feature Fusion Net (PFFNet). The input images first passed through a backbone network for multi-resolution feature maps. The backbone is omitted for brevity. The overall loss function for training is shown in Eq.\ref{loss-equ}. }
\label{overall-fig}
\end{figure*}

In this section, we firstly introduce the overall architecture of the proposed Panoptic Feature Fusion Net (PFFNet). Then, the three newly proposed modules are described in detail. Finally, the training and inference details are presented.

\subsection{Overall Architecture}

\begin{figure*}[!t]
\centering
\includegraphics[width=0.75\textwidth]{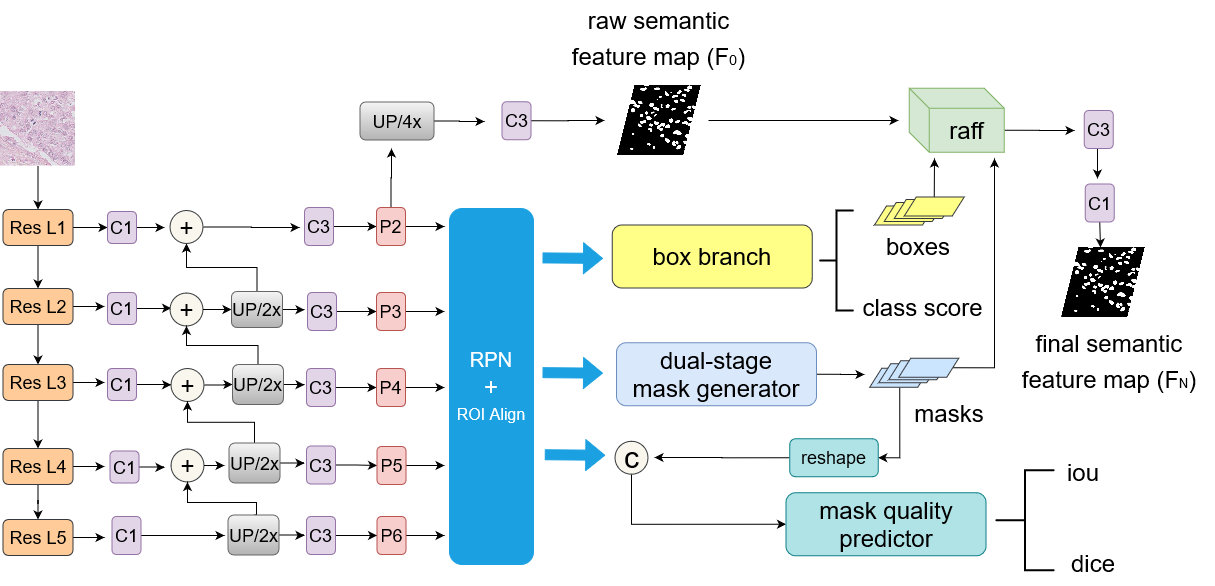}
\caption{Overview of our instance segmentation branch. $C1$ and $C3$ represent the convolutional layer with the kernel size of $1$ and $3$ respectively. The $C$ before the mask quality predictor is the concatenation operation. $UP/nx$ means the upsampling layer for $n$ times with the nearest interpolation. $raff$ represents the proposed residual attention feature fusion mechanism. The ReLU and group normalization layer after all the convolutional layers are omitted for brevity.}
\label{ins-fig}
\end{figure*}

Fig.~\ref{overall-fig} illustrates our proposed PFFNet. For each input image, it first passes through a ResNet-101 \cite{he2016deep} backbone network to obtain the feature maps at different resolutions. Then, the feature maps are sent to a semantic segmentation branch to learn the global semantic-level feature and an instance segmentation branch to learn the object-level local features. 

\begin{figure}[!t]
\centering
\includegraphics[width=0.48\textwidth]{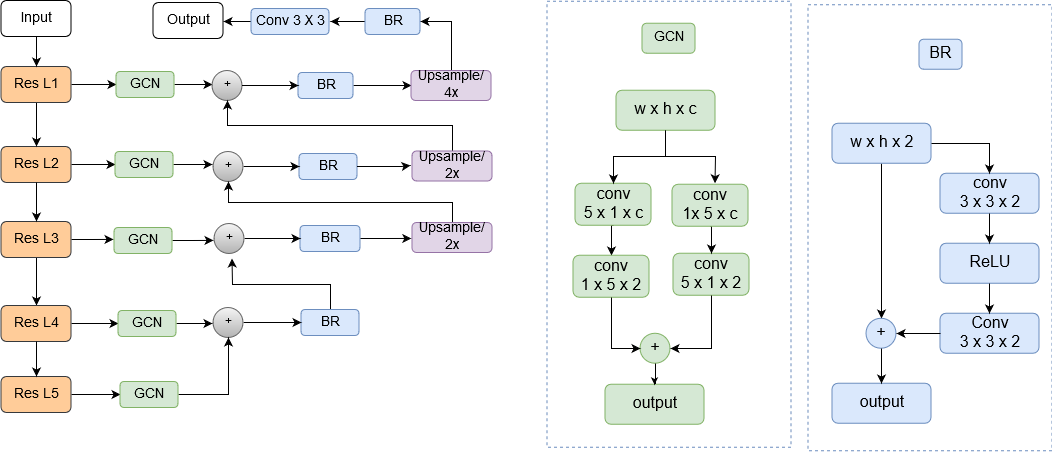}
\caption{Overview of the semantic branch in our proposed PFFNet. }
\label{sem-fig}
\end{figure}

For the semantic segmentation branch, we employed the decoder of the global convolutional network (GCN) \cite{peng2017large}, as shown in Fig.~\ref{sem-fig}. Specifically, multi-resolution feature maps after the ResNet101 backbone network are sent to a skip connected decoder, which contains several large kernel global convolutional modules. Each large kernel global convolutional module is simulated by incorporating two 1D convolutional kernels in different orders. To this end, the model has a large receptive field as well as memory efficiency, and the semantic branch is capable of processing more global-level contextual features compared with the CNN architectures with a normal size convolutional kernels. 

Our instance segmentation branch in Fig.~\ref{ins-fig} is based on that of Cell R-CNN V2 \cite{liu2019nuclei}. First, multi-resolution feature maps ($P2, P3, P4, P5,$ and $P6$ in Fig.~\ref{ins-fig}) are obtained by the feature pyramid network (FPN) \cite{lin2017feature} connected after the backbone encoder. Along with the anchors in different ratios and sizes, $P2, P3, P4, P5,$ and $P6$ then pass through a region proposal network (RPN) \cite{ren2015faster} to generate ROIs which represent the features of all possible foreground objects in the original images. As the ROIs after RPN are in various sizes, a ROIAlign mechanism \cite{he2017mask} is further employed to reshape all the ROIs to the same size, which is $14 \times 14$ in this work. Eventually, all the ROIs are sent to a bounding box sub-branch to predict the locations and class scores and a dual-model mask generator \cite{liu2019nuclei} for mask instance segmentation prediction. In order to induce the semantic feature learning in the decoder of the instance segmentation branch, we further propose an attention-based feature fusion mechanism to incorporate the mask prediction and bounding box prediction for all the ROIs with the semantic segmentation feature map obtained from the top layer of FPN ($P2$). In addition, the mask segmentation result is fused with the ROI features for a newly proposed mask quality sub-branch to predict the quality of the mask segmentation for each ROI according to the corresponding IoU and Dice score.

\subsection{Residual Attention Feature Fusion Mechanism}

\begin{figure*}[!t]
\centering
\includegraphics[width=0.8\textwidth]{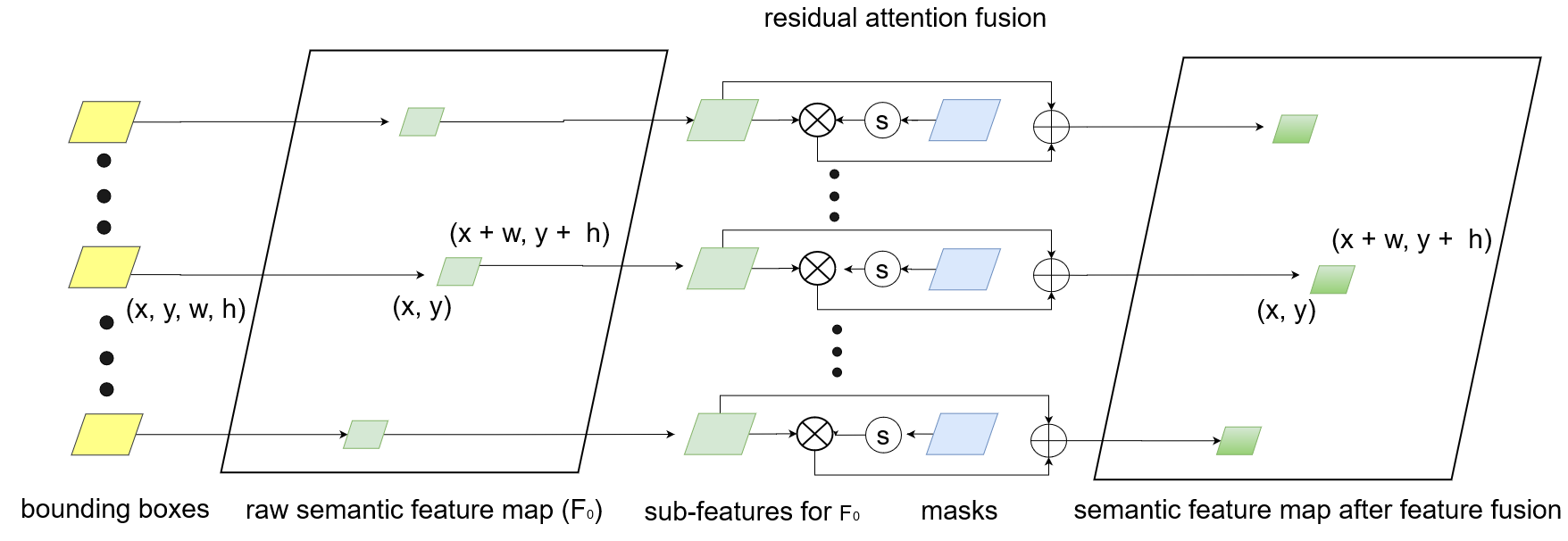}
\caption{The proposed residual attention feature fusion mechanism. $S$ is the sigmoid operation, $\times$ is the element-wise multiplication, and $+$ is the element-wise summation. }
\label{aff-fig}
\end{figure*}

In Cell R-CNN V2, we proposed a feature fusion mechanism to incorporate the semantic-level contextual features with the local-level instance features by using the mask prediction from the instance branch to replace the subset of the semantic segmentation features according to the location of the bounding box sub-branch. Although the fused feature map contains both semantic- and instance-level features, only the background features at the semantic level are learned by the instance segmentation branch, as the foreground features in the original semantic feature map are deprecated. However, the foreground feature for each object from the global view in the original semantic feature maps is still important, as it contains the relationship between each object and the whole background. In the instance segmentation branch, the mask prediction of each object is predicted according to the relationship between the foreground and the background within the corresponding $28 \times 28$ ROI, instead of the background of the whole image. Moreover, part of the background feature in the semantic feature map is also replaced by that from the instance predictions after the feature fusion mechanism in the Cell R-CNN V2, which results in the contextual information loss in the semantic segmentation prediction. To this end, directly replacing the subset of the semantic feature map with the predictions of the instance branch is harmful to the semantic-level feature learning in the decoder of the instance branch. 

To tackle this issue, we design an attention-based feature fusion mechanism, as illustrated in Fig.~\ref{aff-fig}. The number of ROIs in the instance segmentation branch is denoted as $K$, and the mask and bounding box predictions for each ROI are defined as $M_{i}$ and $B_{i}$, respectively, where $i \in [1, K]$. Specifically, $B_{i}$ can be written as:

\begin{equation}
\begin{aligned}
B_{i} = (x_{i}, y_{i}, w_{i}, h_{i})
\end{aligned}
\end{equation}
where $x_i$ and $y_i$ represent the corrdinates of the bottom left point of the $ith$ rectangle ROI in $x$ and $y$ axes and $ w_{i}$ and $ h_{i}$ are its width and height. In addition, the semantic feature map before the attention-based feature fusion is defined as $F_{0}$, as illustrated in Fig.~\ref{ins-fig}. During the attention-based feature fusion for each $M_{i}$, first we obtain its probability map $P_{i}$:

\begin{equation}
\begin{aligned}
P_{i} = \sigma(M_{i})
\end{aligned}
\end{equation}
where $\sigma()$ is the sigmoid operation. Then, we fuse each $P_{i}$ with the subset of $F_{0}$ according to the correpsonding $B_{i}$. as shown in Algorithm~\ref{raff-alg}, where $R(P_{i}, (w_{i}, h_{i}))$ reshapes the $P_{i}$ to $(w_{i}, h_{i})$ with bilinear interpolation, and $*$ is the element-wise multiplication. 

The value of each coordinate of $P_{i}$ represents the probability of this pixel being the foreground. Therefore, the proposed residual attention feature fusion mechanism highlights the foreground features on the original semantic feature map. By fusing the instance-level features on the semantic feature map while preserving all its contextual features, optimizing the semantic segmentation task of the instance branch enables the mask generator to learn accurate and sufficient semantic features.

\begin{algorithm}
\caption{Algorithm for the residual attention feature fusion mechanism}
\label{raff-alg}
\begin{algorithmic}[1]
\REQUIRE ~~\\
Mask probability predictions in the instance branch $P_{i}$, its corresponding bounding box $B_{i}$, and the raw semantic feature $F_{0}$. $i = 1,\dots,K$
\FOR{$i \in [1,K]$}
\STATE $sub(F_{0}, B_{i}) =  F_{i-1}[x_{i}: x_{i} + w_{i}, y_{i}: y_{i} + h_{i}];$ \
\STATE $F_{i} =  sub(F_{0}, B_{i}) * (1 + R(P_{i}, (w_{i}, h_{i})));$ \
\ENDFOR
\RETURN $F_{K}$
\end{algorithmic}
\end{algorithm}

\subsection{Mask Quality Sub-branch}


During the inference process of the traditional Mask R-CNN, the mask predictions are determined by the highest classification score. However, the classification scores for the mask predictions are not always correlated with their quality, such as the IoU between the mask prediction and the ground truth \cite{huang2019mask}. In the testing phase of the Cell R-CNN and Cell R-CNN V2, if there remain two overlapping predictions, the overlapping part is assigned to the mask with a higher classification score. Therefore, low-quality mask predictions with high classification scores affect the performance when processing the overlapping objects during inference.

Inspired by \cite{huang2019mask}, we propose a new mask quality sub-branch to predict the quality of the mask predictions in the instance branch, as shown in Fig. \ref{ins-fig}. For each mask prediction in size $2 \times 28 \times 28$, we select its foreground $ 1 \times 28 \times 28$ score map. Then, each $ 1 \times 28 \times 28$ score map is reshaped to size $ 4 \times 14 \times 14$, to concatenate with the $256 \times 14 \times 14$ ROI feature map. The fused feature map with size $260 \times 14 \times 14$ then passes through $3$ convolutional layers and $3$ fully connected layers to predict the quality of the mask, which is a float value in $(0,1)$. Table \ref{maskqua-table} indicates the detailed hyperparameters setting in the mask quality sub-branch.
As Dice coefficient is an important metric in the biomedical and biological images segmentation task, the mask quality score is determined by the IoU and Dice score between the predictions and the ground truth.
During training, a mask prediction and its corresponding ground truth are denoted as $M_{p}$ and $M_{t}$, respectively, the mask quality score $s_{qua}$ is defined as:

\begin{equation}
\begin{aligned}
s_{qua} = (2 * \frac{|M_{p} \cap M_{t}|}{|M_{p}| + |M_{t}|} + \frac{|M_{p} \cap M_{t}|}{|M_{p} \cup M_{t}|}) * 0.5
\end{aligned}
\end{equation}
where $|.|$ means the total number of the pixels. Therefore, $s_{qua}$ is in $(0,1)$. Eventually, $l2$ loss is employed between the $s_{qua}$ and the mask quality prediction.

\begin{table}[ht]
\centering
\caption{The parameters for each block in our proposed mask generator. $k$, $s$, and $p$ denote the kernel size, stride, and padding of the convolution operation, respectively.}
\resizebox{0.75\linewidth}{!}{%
\begin{tabular}{l|l|l}
\hline
Stage  & Hyperparamaters & Output size  \\ 
\hline
Input            &  & $260 \times 14 \times 14$   \\ \hline
 Conv1             & $k = (3,3), s = 1, p = 1$ & $256 \times 14 \times 14$   \\ \hline
 Conv2             & $k = (3,3), s = 1, p = 1$ & $256 \times 14 \times 14$   \\ \hline
 Conv3             & $k = (3,3), s = 1, p = 1$ & $256 \times 14 \times 14$   \\ \hline
 Conv4             & $k = (3,3), s = 2, p = 1$ & $256 \times 7 \times 7$   \\ \hline
 FC1        &  & $1 \times 1 \times 1024$   \\ \hline
 FC2        &  & $1 \times 1 \times 1024$   \\ \hline
 FC3        &  & $1 \times 1 \times 1$   \\ \hline
  Output   &  & $1 \times 1 \times 1$   \\ \hline
\end{tabular}}
\label{maskqua-table}
\end{table}

\subsection{Semantic Task Consistency Regularization}

Our motivation for this module is from \cite{chen2018domain}. When there are two tasks in a multi-task learning architecture focusing on the same objective, adding a consistency regularization between the outputs of these two tasks enables the robust learning of both. In our proposed architecture, both the semantic and the instance branches generate semantic segmentation predictions. In the ideal situation, the semantic segmentation predictions from both two branches should be equal to each other and equal to the ground truth. Therefore, we propose a consistency regularization between these two semantic segmentation predictions to reduce the distance between them. The softmax semantic segmentation prediction of the semantic and instance branch are denoted as $p_{sem}$ and $p_{ins}$, respectively, which are both in range $(0,1)$. The semantic consistency regularization is:

\begin{equation}
L_{sem-cons} = \frac{1}{N}\sum_{i,j}(p_{sem (i,j)} - p_{ins (i,j)})^{2}
\label{cons-equ}
\end{equation}
where $N$ is the total number of activations in the $p_{sem}$.

\subsection{Training and Inference Details}

As shown in Fig. \ref{overall-fig}, the total loss function of the PFFNet is defined as:

\begin{equation}
\label{loss-equ}
\begin{aligned}
    L_{overall} & = L_{rpn-obj} + L_{rpn-reg} + L_{det-cls} \\
                 &  + L_{det-reg} + L_{det-mask} +  L_{det-qua}\\
                 & + \alpha_{1} (L_{semseg1} + L_{semseg2}) + \alpha_{2} L_{sem-cons}
\end{aligned}
\end{equation}
For the instance segmentation task, $ L_{rpn-obj}$ and $ L_{rpn-cls}$ are the smooth L1 regression loss and cross entropy classification loss for RPN, respectively. $L_{det-reg}$ and $L_{det-cls}$ are the bounding box regression and the classification loss of the box sub-branch, $ L_{det-mask} $ is the binary cross entropy segmentation loss for the mask sub-branch, and $ L_{det-qua}$ is the $l2$ regression loss for the mask quality sub-branch. On the other hand, $L_{semseg1}$ and $L_{semseg2}$ are the semantic segmentation losses for the semantic branch and instance branch. $ L_{sem-cons}$ is the mean square loss for the semantic consistency regularization, as shown in Eq. \ref{cons-equ}. $\alpha_{1} $ and $\alpha_{2}$ are trade-off parameters to balance the importance of each task and are set as $0.1$ and $1$, respectively, in our experiments.

During inference, the instance mask predictions from the mask generator of the instance branch are employed. A confident threshold score $\beta$ is firstly employed to depreciate the masks whose classification scores are smaller than $\beta$. Then, a mask confidence score $s_{conf}$ for each object is calculated based on its classification score $s_{cls}$ and mask quality prediction $s_{qua}$:

\begin{equation}
\begin{aligned}
s_{conf} = s_{cls} * s_{qua}
\end{aligned}
\end{equation}
For any two touching predictions, the overlapping part belongs to the prediction with the higher $s_{conf}$. 

\section{Experiments}

\subsection{Dataset Description}

\subsubsection{TCGA-KUMAR}

This dataset contains $30$ histopathology images in size $1000 \times 1000$, obtained from the The Cancer Genome Atlas (TCGA) at $40 \times$ magnification \cite{kumar2017dataset}. Each image is from one of the seven organs, including breast, bladder, colon, kidney, liver, prostate, and stomach. In order to compare with the state-of-the-art methods, we have the same data split as in \cite{kumar2017dataset,naylor2018segmentation,liu2019nuclei}. $12$ images total from the breast, kidney, liver, and prostate are employed for training ($3$ from each organ). During training, $20$ patches in size $256 \times 256$ are randomly cropped from each $1000 \times 1000$ image. Next, basic augmentation techniques are applied, including horizontal and vertical flipping and rotation of  $90^{\circ}$, $180^{\circ}$, and $270^{\circ}$. Due to the noise and variability of color in the histopathology images, advanced augmentation including Gaussian blur, median blur, Gaussian noise are then employed to ensure the robustness of the model. The validation set contains $4$ images from the breast, kidney, liver, and prostate. For the remaining $14$ images, $8$ images from the same $4$ organs in the training set form the seen testing set, while $6$ from the other $3$ organs unavailable to the training are selected as the unseen testing set. During testing, each $1000 \times 1000$ image is directly employed for nuclei instance segmentation.

\subsubsection{TNBC}

This is our second histopathology dataset focusing on the Triple Negative Breast Cancer (TNBC) dataset from \cite{naylor2018segmentation}. The TNBC dataset contains $30$ $512 \times 512$ histopathology images at $40 \times$ magnification, collected from $11$ different patients of the Curie Institute. We conduct 3-fold cross validation for all the experiments on this dataset. During training, $5$ $256 \times 256$ patches are cropped from each $512 \times 512$ images, following data augmentation including including horizontal and vertical flipping, rotation of $90^{\circ}$, $180^{\circ}$, and $270^{\circ}$, Gaussian blur, median blur, and Gaussian noise. For testing, each $512 \times 512$ image is directly employed.

\subsubsection{Fluorescence microscopy images}

In addition to the histopathology images, we also validate our PFFNet on the fluorescence microscopy images analysis. We employ the BBBC039V1 dataset from \cite{ljosa2012annotated}, which contains $200$ $520 \times 696$ images obtained from fluorescence microscopy. Each image focuses on the U2OS cells with a single field of view on the DNA channel, with various cell shape and density. In our experiment, we follow the official data split ({https://data.broadinstitute.org/bbbc/BBBC039/}), with $100$ images for training, $50$ for validation, and the rest $50$ for testing. For training data preparation, first, $10$ $256 \times 256$ patches are randomly cropped from each image. As the background components in this dataset are not as complicated as the others, only basic data augmentation is employed, including horizontal and vertical flipping and rotation of $90^{\circ}$, $180^{\circ}$, and $270^{\circ}$. During inference, each $520 \times 696$ image is directly used.

\subsubsection{Plant phenotyping}

To demonstrate the effectiveness of our proposed PFFNet on instance segmentation task for other biology images, we study the leaf instance segmentation task. We employ the Computer Vision Problems in Plants Phenotyping (CVPPP) \cite{minervini2016finely} dataset, which contains top-down view images of leaves with various shapes and complicated occlusions. In this work, we focus on the A1 subset with a total of $161$ $ 530 \times 500$ images, which has been broadly studied for instance segmentation in several state-of-the-art works. Out of the $128$ training images provided by the challenge, we employed $100$ images for training and the remaining $28$ for validation. During training, each image is firstly reshaped to size $512 \times 512$. Then, data augmentation including horizontal and vertical flipping, rotation of  $90^{\circ}$, $180^{\circ}$, and $270^{\circ}$, Gaussian blur, median blur, and Gaussian noise are employed to avoid overfitting. During inference, the predictions are directly obtained from the $530 \times 500$ images. To evaluate the performance, the predicted results are submitted to the official evaluation platform ({https://competitions.codalab.org/competitions/18405}).

\subsection{Evaluation Metrics}

To evaluate the performance on the nuclei segmentation in the histopathology images and cell segmentation in the fluorescence microscopy images, we employed Aggregated Jaccard Index ($AJI$), object-level F1 score ($F1$), Panoptic Quality ($PQ$), and pixel-level Dice score ($Dice$). $AJI$ is an extended Jaccard Index for object-level segmentation evaluation \cite{kumar2017dataset}, defined as:

\begin{equation}
AJI = \frac{\sum_{i = 1}^{N}|G_{i} \cap P_{M}^{i}|}{\sum_{i = 1}^{N} |G_{i} \cup P_{M}^{i}| + \sum_{F \in U}|P_{F}|}
\end{equation}
where $G_{i}$ is the $ith$ nucleus in a ground truth with a total of $N$ nuclei. U is the set of false positive predictions without the corresponding ground truth. For each ground truth object $G_{i}$, $M$ is the index of the prediction with the largest overlapping with it and each $M$ can only be used once, which is defined as:

\begin{equation}
M = argmax \frac{P_{M}^{i} \cap G_{i}}{P_{M}^{i} \cup G_{i}}
\end{equation}
Object-level F1 score is the metric for the detection performance~\cite{chen2017dcan}, defined based on the number of true and false detections: 

\begin{equation}
\begin{aligned}
F1 = \frac{2TP}{FN + 2TP + FP}
\end{aligned}
\end{equation} 
where TP, FN, and FP represent the number of true positive (corrected detected objects), false negative (ignored objects), and false positive (detected objects without corresponding ground truth) detections, respectively. Note that a true positive object for object-level F1 score should intersect with more than $50\%$ of its corresponding ground truth. Panoptic Quality (PQ) has been previously employed to evaluate the performance of the panoptic segmentation tasks~\cite{kirillov2018panoptic,kirillov2019panoptic}, which is the multiplication between the Detection Quality (DQ) for object detection, and Segmentation Quality (SQ) for object segmentation. PQ is defined as: 
\begin{equation}
\label{equ-pq}
\begin{aligned}
PQ = \underbrace{\frac{2|TP|}{2|TP| + |FP| + |FN|}}_{DQ} \times \underbrace{\frac{\sum_{(p, g) \in TP} IoU(p, g)}{|TP|}}_{SQ}
\end{aligned}
\end{equation} 
where $|TP|$, $|FN|$, and $|FP|$ represent the number of true positive, false negative, and false positive detections, respectively. Each $(p, g)$ indicates a pair of mask prediction from the true positive detections, and its corresponding ground truth. Note that a mask prediction can only be regarded as the true positive when $IoU(p, q) > 0.5$. As illustrated in Eq.~\ref{equ-pq}, the $PQ$ metric reflects the performance on object detection and segmentation. To evaluate the foreground and background segmentation accuracy, pixel-level Dice score is employed between the binarized prediction and the ground truth:

\begin{equation}
\begin{aligned}
Dice = \frac{2|P \cap G|}{|P| + |G|}
\end{aligned}
\end{equation} 
where $P$ and $G$ represent the binarization prediction and ground truth, respectively. $|.|$ means the total number of foreground pixels.

For the evaluation metrics of the leaf segmentation task, we directly employ the official Symmetric Best Dice ($SBD$) score:

\begin{equation}
\begin{aligned}
SBD(P, T) =  min (BD(P, T), BD(T, P))
\end{aligned}
\end{equation}
where $P$ and $T$ are the predictions and ground truth, respectively. $BD(P, T)$ is the best dice between ${P_{i}}_(i = 1,\dots,M)$ and ${T_{j}}_(j = 1,\dots,N)$:
\begin{equation}
\begin{aligned}
BD(P, T) = \frac{1}{M}\sum_{i=1}^{M}\max_{j = 1,\dots,N}\frac{2|P_i \cap T_j|}{|P_i| + |T_j|}
\end{aligned}
\end{equation}
where $|.|$ means the total number of foreground pixels. 

\subsection{Implementation Details}

For the network initialization, the weights of the ResNet101 backbone are pretrained on the ImageNet \cite{deng2009imagenet} classification task, while the weights for other layers are initialized with ``Kaiming" initialization \cite{he2015delving}. When training the PFFNet, stochastic gradient descent (SGD) is used to optimize the network, with a weight decay of $0.0001$, and momentum of $0.9$. The mini-batch size is $1$, which is relatively a small batch size. We, therefore, employed group normalization layers \cite{wu2018group} with a group number of $32$ to replace the traditional batch normalization layers. The initial learning rate is set to $0.003$, with a linear warm-up for the first $500$ iterations. The learning rate is then decreased to $0.0003$ when it reaches the $3/4$ of the total training iterations. Our experiments are implemented on two Nvidia GeForce 1080Ti GPUs with Pytorch \cite{paszke2017automatic}.

\subsection{Comparison with State-of-the-art Instance Segmentation Methods for Biomedical and Biological Images~\label{sec-cmp-bio}}

\subsubsection{TCGA-KUMAR}

\begin{table*}[!t]
\centering
\caption{The comparison of results for TCGA-Kumar dataset. avg and std represent average and standard deviation, respectively. For DIST, the results of object-level $F1$ and $PQ$ are unknown. For CNN3, the $PQ$ score is unknown.
}
\resizebox{0.99\linewidth}{!}{%
\begin{tabular}{c|c|c|c|c|c|c|c|c|c|c|c|c|c}
\hline
\multicolumn{2}{c|}{\multirow{2}{*}{Methods}} &  \multicolumn{3}{c|}{$AJI$}  &  \multicolumn{3}{c|}{$Dice$}  &  \multicolumn{3}{c|}{$F1$} &  \multicolumn{3}{c}{$PQ$}\\
\cline{3-14}
\multicolumn{2}{c|}{} & seen & unseen  & all & seen & unseen  & all  & seen & unseen  & all & seen & unseen  & all\\
\cline{1-14}
\multirow{2}{*}{CNN3 \cite{kumar2017dataset}}  & avg  &$0.5154 $ &  $0.4989$ &  $0.5083$ &  $0.7301 $ &  $0.8051 $ &  $0.7623$ &  $0.8226 $ &  $0.8322 $ &  $0.8267$ &  $- $ &  $- $ &  $-$\\
\cline{2-14}
 & std  &$0.0835 $ &  $0.0806$ &  $0.0695$ &  $0.0590 $ &  $0.1006 $ &  $0.0946$ &  $0.0853 $ &  $0.0764 $ &  $0.0934$ &  $- $ &  $- $ &  $-$\\
\hline
\multirow{2}{*}{DIST \cite{naylor2018segmentation}}  & avg  &$0.5594 $ &  $0.5604$ &  $0.5598$ &  $0.7756 $ &  $0.8005 $ &  $0.7863$ &  $- $ &  $- $ &  $-$ &  $- $ &  $- $ &  $-$\\
\cline{2-14}
 & std &$0.0598 $ &  $0.0663$ &  $0.0781$ &  $0.0489 $ &  $0.0538$ &  $0.0550$ &  $- $ &  $- $ &  $-$ &  $- $ &  $- $ &  $-$\\
 \hline
\multirow{2}{*}{Mask R-CNN \cite{he2017mask}}  & avg  &$0.5438$ &  $0.5340$ &  $0.5396$ &  $0.7659 $ &  $0.7658 $ &  $0.7659$ &  $0.6987 $ &  $0.6434$ &  $0.6750$ &  $0.4856 $ &  $0.4715 $ &  $0.4796$\\
\cline{2-14}
  & std  &$0.0649 $ &  $0.1283$ &  $0.0929$ &  $0.0481 $ &  $0.0608 $ &  $0.0517$ &  $0.1344$ &  $0.1908$ &  $0.1566$&  $0.0893 $ &  $0.1709 $ &  $0.1248$\\
 \hline
 \multirow{2}{*}{Cell R-CNN \cite{zhang2018panoptic}}  & avg  &$0.5547 $ &  $0.5606$ &  $0.5572$ &  $0.7746 $ &  $0.7752 $ &  $0.7748$&  $0.7587 $ &  $0.7481 $ &  $0.7542$&  $0.5066 $ &  $0.5098 $ &  $0.5079$\\
\cline{2-14}
  & std  &$0.0567 $ &  $0.1100$ &  $0.0800$ &  $0.0446 $ &  $0.0577 $ &  $0.0485$&  $0.0969 $ &  $0.1488$ &  $0.1166$&  $0.0816 $ &  $0.1392 $ &  $0.1051$\\
\hline
\multirow{2}{*}{Cell R-CNN V2 \cite{liu2019nuclei}}  & avg  &$0.5758 $ &  $0.5999$ &  $0.5861$ &  $0.7841 $ &  $0.8078$ &  $0.7943$&  $0.8014 $ &  $0.8023 $ &  $0.8017$&  $0.5500 $ &  $0.5563 $ &  $0.5527$\\
\cline{2-14}
 & std  &$0.0568 $ &  $0.1160$ &  $0.0841$ &  $0.0439 $ &  $0.0611 $ &  $0.0512$&  $0.0757 $ &  $0.1081 $ &  $0.0871$&  $0.0748 $ &  $0.1346 $ &  $0.1000$\\
 \hline
 \hline
\multirow{2}{*}{PFFNet}  & avg  &$ \textbf{0.5975} $ &  $ \textbf{0.6282}$ &  $ \textbf{0.6107}$ &  $ \textbf{0.7967}$ &  $ \textbf{0.8256} $ &  $ \textbf{0.8091}$&  $ \textbf{0.8317} $ &  $ \textbf{0.8383} $ &  $ \textbf{0.8345}$ &  $ \textbf{0.5824} $ &  $ \textbf{0.5933} $ &  $ \textbf{0.5871}$\\
\cline{2-14}
 & std  &$0.0568 $ &  $0.0924$ &  $0.0726$ &  $0.0453 $ &  $0.0520 $ &  $0.0487$&  $0.0694 $ &  $0.0598 $ &  $0.0631$&  $0.0767 $ &  $0.1023 $ &  $0.0850$\\
 \hline
\end{tabular}}
\label{tcga-sup}
\end{table*}

\begin{figure*}[!t]
\centering
\includegraphics[width=0.72\textwidth]{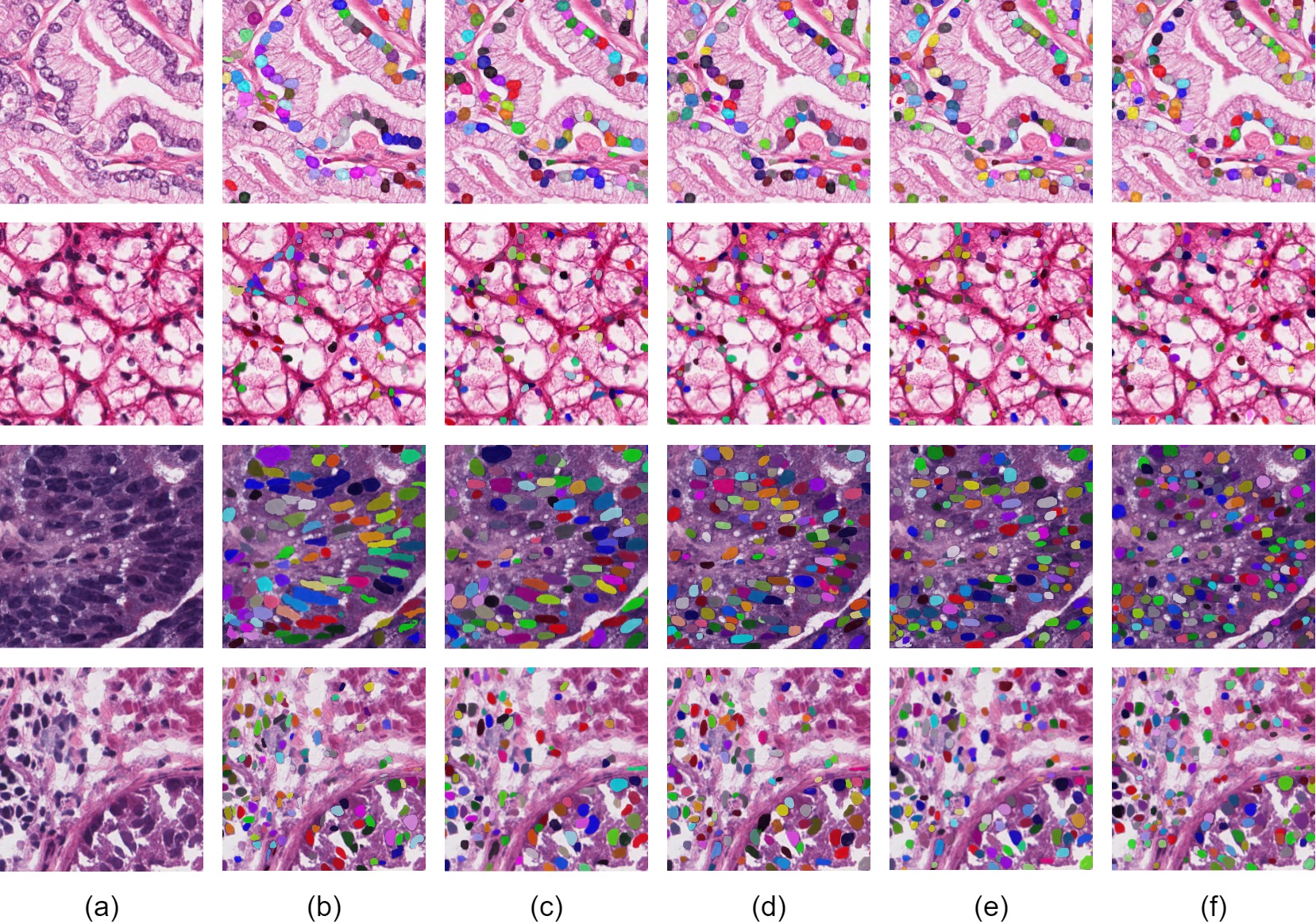}
\caption{The visual comparison of results for TCGA KUMAR dataset. (a) original images, (b) ground truth annotations, (c) predictions by our proposed PFFNet, (d) predictions by Cell R-CNN V2 \cite{liu2019nuclei}, (e) predictions by Cell R-CNN \cite{zhang2018panoptic}, and (f) predictions by Mask R-CNN \cite{he2017mask}. }
\label{cmp-vis-tcga}
\end{figure*}

\begin{figure*}[!htb]
\centering
\includegraphics[width=0.98\linewidth]{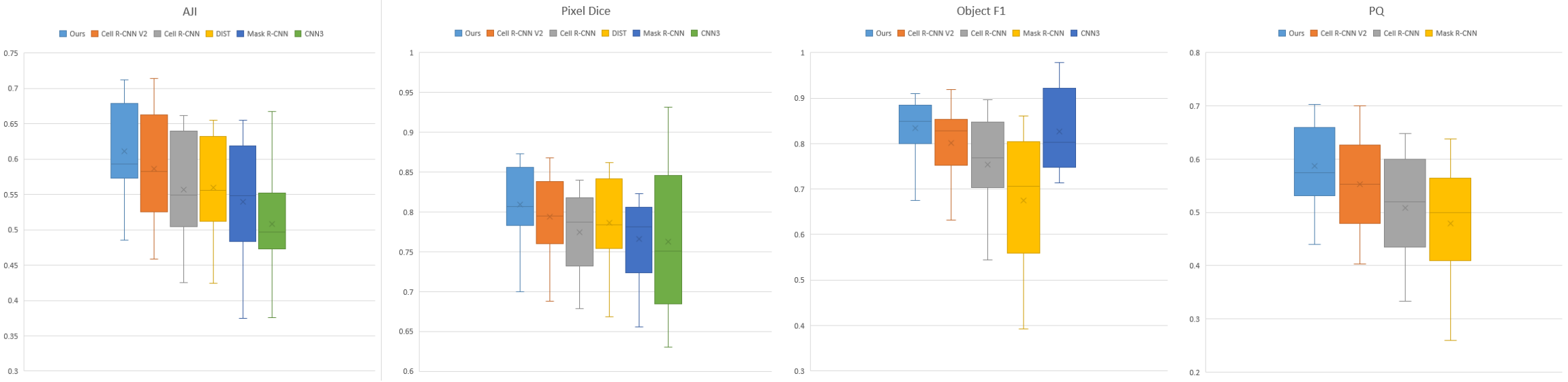}
\caption{Box plot for all the compared methods on TCGA KUMAR dataset. The object-level F1 and PQ score and were not available in the original DIST work. The PQ score was not reported by the original CNN3 work.}
\label{bp-tcga}
\end{figure*}

Our result is compared with several state-of-the-art nuclei instance segmentation methods, including CNN3 \cite{kumar2017dataset}, DIST \cite{naylor2018segmentation}, Mask R-CNN \cite{he2017mask}, Cell R-CNN \cite{zhang2018panoptic}, and Cell R-CNN V2 \cite{liu2019nuclei}. With the same data split, we directly compare the performance reported in \cite{kumar2017dataset,naylor2018segmentation}. For Mask R-CNN, Cell R-CNN, and Cell R-CNN V2, we re-implement them by adding group normalization with the same settings as our proposed PFFNet, for a fair comparison. Therefore their performance is slightly better than in \cite{liu2019nuclei}. Table \ref{tcga-sup} and Fig. \ref{cmp-vis-tcga} illustrate our quantitative and qualitative comparison results, respectively. As shown in Table \ref{tcga-sup}, our proposed PFFNet outperforms all the other methods in all four metrics on the seen and unseen testing set. It indicates that our PFFNet has a strong generalization ability when testing on the cases from the unseen organs. In order to test the statistical significance between the results of our PFFNet and other methods, we employed one-tailed-paired t-test to calculate the p-value. As shown in Table \ref{p-value-tcga}, our improvements under all four metrics is statistically significant (p-value $ < 0.05$) except for the $F1$ of CNN3. However, $F1$ only relies on the number of corrected detected objects, regardless of the segmentation quality of each detected object. By outperforming CNN3 by a large margin in the other two segmentation metrics (over $10 \%$ on $AJI$ and $4 \%$ on $Dice$), our PFFNet still achieves better performance on nuclei segmentation tasks compared with CNN3. Fig.~\ref{bp-tcga} is the box plot for all the compared method under the four metrics, which shows that our proposed PFFNet not only outperforms all the methods, but is also more stable and robust.

\begin{table}[!t]
\caption{\label{p-value-tcga}p-value for the methods in Table \ref{tcga-sup} compared with our proposed PFFNet, on TCGA KUMAR dataset under all the four metrics.}
\centering
\resizebox{0.99\linewidth}{!}{%
\begin{tabular}{l|l|l|l|l}
\hline
 & $AJI$ & $Dice$ & $F1$  & $PQ$ \\
 \hline
CNN3  & $2.69 \times 10^{-3}$ & $1.67 \times 10^{-2}$ & $0.34$ & $-$\\
\hline
DIST & $2.35 \times 10^{-6}$ & $1.27 \times 10^{-5}$ & $-$ & $-$\\
\hline
Mask R-CNN & $6.44 \times 10^{-5}$ & $1.48 \times 10^{-6}$ & $5.57 \times 10^{-5}$ & $1.14 \times 10^{-5}$ \\
\hline
Cell R-CNN & $3.93 \times 10^{-6}$ & $2.39 \times 10^{-6}$ & $2.53 \times 10^{-4}$ & $6.14 \times 10^{-7}$ \\
\hline
Cell R-CNN V2 & $7.02 \times 10^{-4}$ & $1.03 \times 10^{-5}$ & $1.33 \times 10^{-3}$& $1.07 \times 10^{-4}$  \\
\hline
\end{tabular}}
\end{table}


\subsubsection{TNBC}

\begin{figure*}[!t]
\centering
\includegraphics[width=0.72\textwidth]{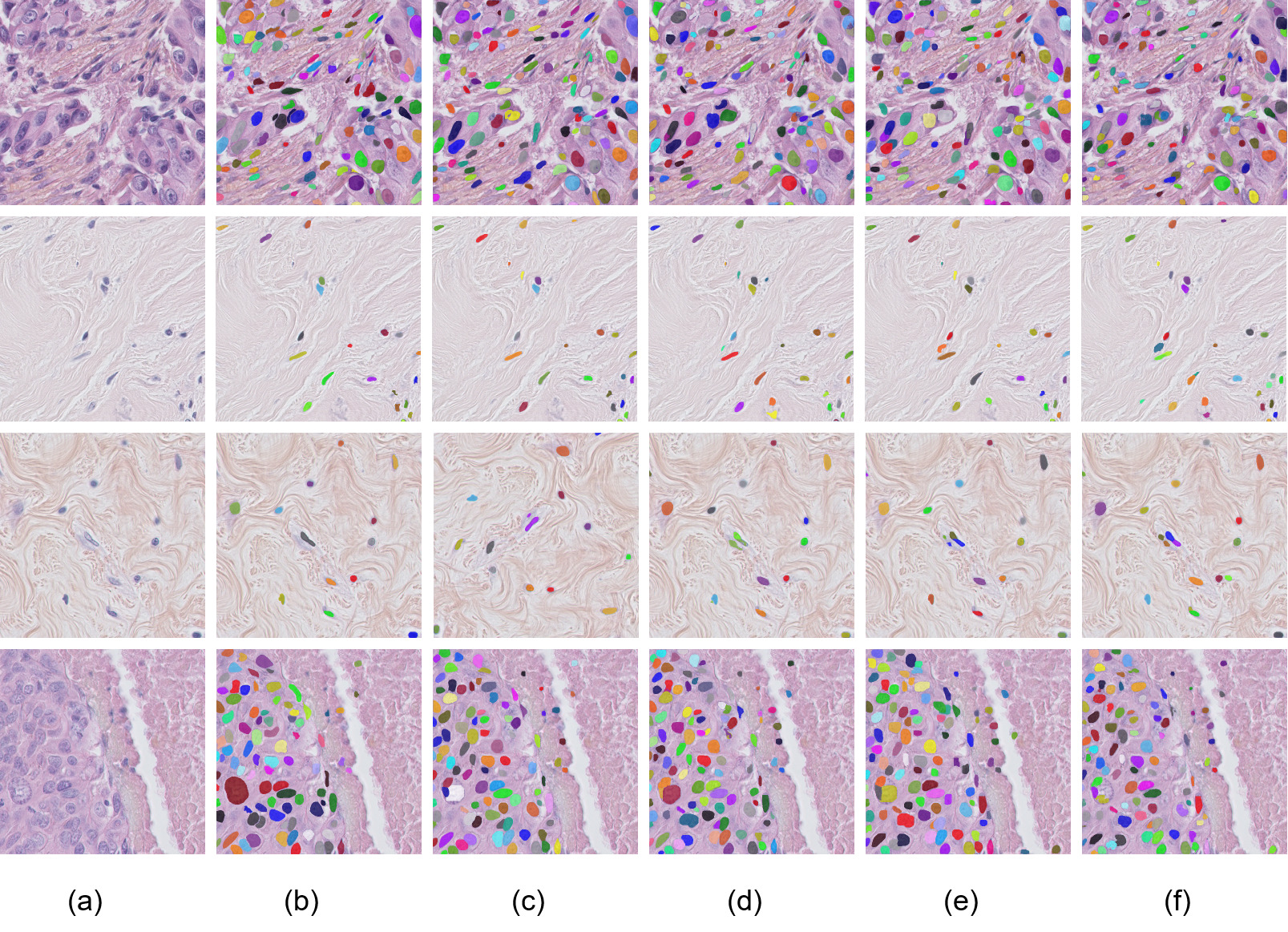}
\caption{The visual comparison of results for TNBC dataset. (a) original images, (b) ground truth annotations, (c) predictions by our proposed PFFNet, (d) predictions by Cell R-CNN V2 \cite{liu2019nuclei}, (e) predictions by Cell R-CNN \cite{zhang2018panoptic}, and (f) predictions by Mask R-CNN \cite{he2017mask}. }
\label{cmp-vis-tnbc}
\end{figure*}

\begin{table*}[!t]
\caption{\label{tnbc-cmp}Comparison experiments on TNBC dataset under $AJI$, $Dice$, $F1$, and $PQ$. Results are presented as mean value with standard deviation in the parentheses.}
\centering
\resizebox{0.75\linewidth}{!}{
\begin{tabular}{l|l|l|l|l}
\hline
Methods & $AJI$ & $Dice$ & $F1$ & $PQ$\\
 \hline
Mask R-CNN & $0.5350 (0.0993)$ & $0.7393 (0.0977)$ & $0.7542 (0.1535)$ & $0.5146 (0.1193)$\\
\hline
Cell R-CNN & $0.5747 (0.1061)$ & $0.7637 (0.1080)$ & $0.8142 (0.1331)$  & $0.5664 (0.1120)$ \\
\hline
Cell R-CNN V2 & $0.5986 (0.0847)$ & $0.7793 (0.0772)$ & $0.8184 (0.1163)$  & $0.5845 (0.0964)$ \\
\hline
PFFNet & $\textbf{0.6313 (0.0750)}$ & $\textbf{0.8037 (0.0557)}$ & $\textbf{0.8600 (0.0849)}$  & $\textbf{0.6298 (0.0820)}$ \\
\hline
\end{tabular}}
\end{table*}

We conducted comparison experiments on the second histopathology dataset with 3-fold cross-validation and the results are shown in Table \ref{tnbc-cmp} and Fig. \ref{cmp-vis-tnbc}. As in Table \ref{tnbc-cmp}, our PFFNet outperforms its previous versions under all three metrics. Compared with Mask R-CNN, the effectiveness of the Cell R-CNN is improved by a large margin. The background components in the TNBC dataset are complicated and some background textures have a similar appearance to the foreground. Therefore, processing the semantic-level information is beneficial to the segmentation and detection accuracies. Compared with the Cell R-CNN, the improvement of Cell R-CNN V2 is not as large as in the TCGA KUMAR dataset, especially under the object-level F1 score. Although the feature fusion mechanism in the Cell R-CNN V2 facilitates the semantic feature learning in the instance branch, there is a lack of contextual features around each object due to the depreciation of part of the semantic feature map. Therefore, the detection accuracies are affected when the boundaries of two touching objects become ambiguous. Similar to the results on the TCGA KUMAR dataset, our proposed PFFNet outperforms the compared methods by a large margin.

\subsubsection{BBBC039V1}

\begin{table*}[!t]
\caption{\label{fluo-cmp}Comparison experiments on BBBC039V1 dataset under $AJI$, $Dice$, $F1$, and $PQ$. Results are presented as mean value with standard deviation in the parentheses.}
\centering
\resizebox{0.75\linewidth}{!}{
\begin{tabular}{l|l|l|l|l}
\hline
Methods & $AJI$ & $Dice$ & $F1$ & $PQ$\\
 \hline
Mask R-CNN & $0.7983 (0.0858)$ & $0.9277 (0.0126)$ & $0.9180 (0.0870)$ & $0.7773 (0.0959)$\\
\hline
Cell R-CNN & $0.8070 (0.0934)$ & $0.9290 (0.0273)$ & $0.9276 (0.0836)$ & $0.7959 (0.0894)$\\
\hline
Cell R-CNN V2 & $0.8260 (0.0779)$ & $0.9336 (0.0097)$ & $0.9328 (0.0728)$ & $0.8010 (0.0839)$\\
\hline
PFFNet & $\textbf{0.8477 (0.0757)}$ & $\textbf{0.9478 (0.0071)}$ & $\textbf{0.9451 (0.0536)}$ & $\textbf{0.8331 (0.0724)}$ \\
\hline
\end{tabular}}
\end{table*}

\begin{figure*}[!t]
\centering
\includegraphics[width=0.72\textwidth]{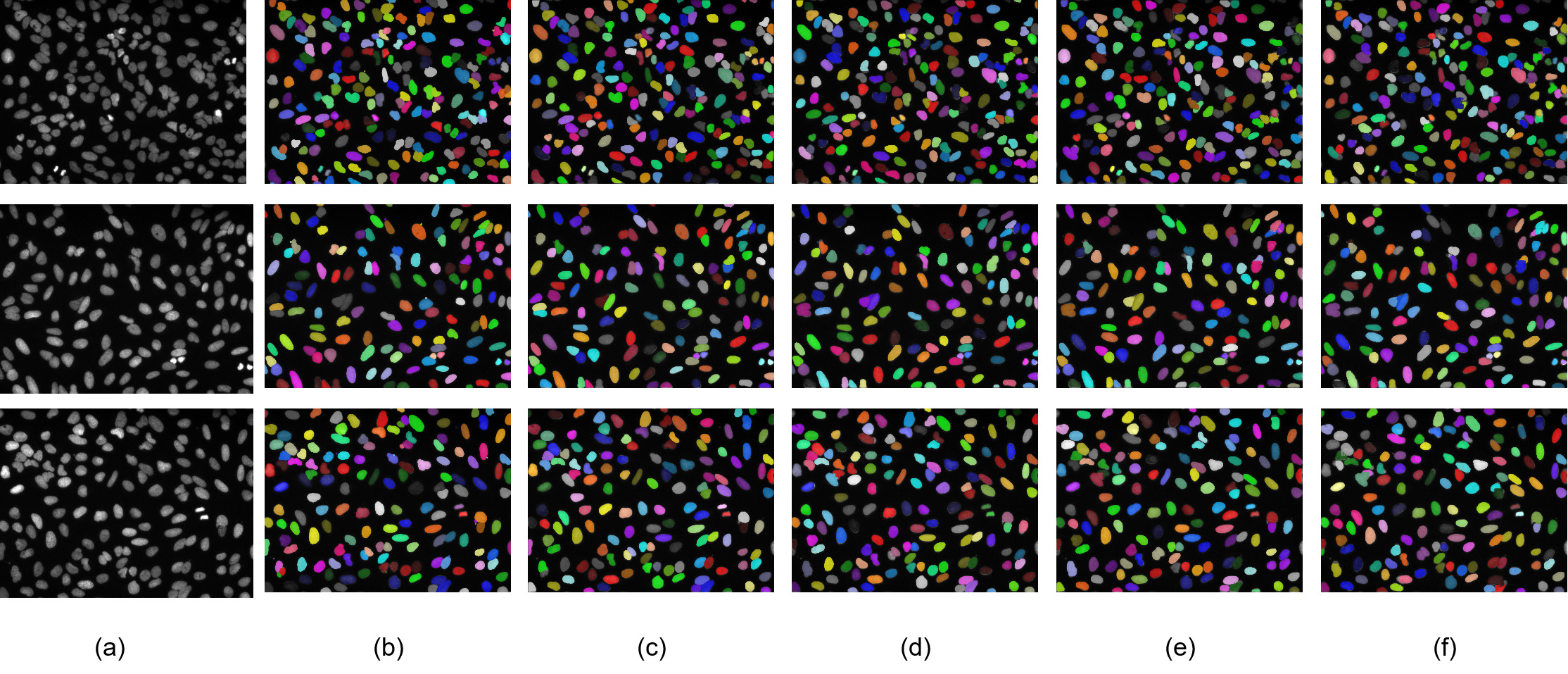}
\caption{The visual comparison of results for BBBC039 V1 dataset. (a) original images, (b) ground truth annotations, (c) predictions by our proposed PFFNet, (d) predictions by Cell R-CNN V2 \cite{liu2019nuclei}, (e) predictions by Cell R-CNN \cite{zhang2018panoptic}, and (f) predictions by Mask R-CNN \cite{he2017mask}. }
\label{cmp-vis-fluo}
\end{figure*}

In addition to the nuclei segmentation tasks in the histopathology images, our proposed PFFNet is also effective for cell instance segmentation in the fluorescence microscopy images. As illustrated in Table \ref{fluo-cmp}, our PFFNet outperforms all the compared methods. We notice that the performance of Cell R-CNN is at the same level as Mask R-CNN, due to the limited improvement. In the fluorescence microscopy images, the background components are not as complicated as in the histopathology images, as shown in Fig. \ref{cmp-vis-fluo}. Therefore, Cell R-CNN is not capable of improving accuracy as it fails to process the contextual information about the foreground objects by learning the semantic features in the backbone encoder. On the other hand, Cell R-CNN V2 improves Cell R-CNN by designing a dual-modal mask generator for improving the mask segmentation accuracies in the instance branch and inducing the mask generator to learn global semantic-level features. However, the improvement of the pixel-level Dice score of Cell R-CNN is still limited. Based on Cell R-CNN V2, our proposed PFFNet achieves high improvements in all four metrics.

\subsubsection{CVPPP Challenge}

\begin{table}[!t]
\caption{\label{cvppp-cv3-cmp} Comparison experiments on CVPPP dataset with our previous methods. Results are presented as mean value with standard deviation in the parentheses.}
\centering
\resizebox{0.98\linewidth}{!}{
\begin{tabular}{l|l|l|l}
\hline
Methods & $SBD$ & $Dice$  & $PQ$\\
 \hline
Mask R-CNN & $0.8674 (0.0362)$ & $0.8453(0.0799)$  & $0.6342(0.0860)$ \\
\hline
Cell R-CNN & $0.8771 (0.0274)$ & $0.8772 (0.0524)$ & $0.6804(0.0750)$ \\
\hline
Cell R-CNN V2 & $0.8861 (0.0259)$ & $0.8931 (0.0434)$ & $0.7161(0.0669)$ \\
\hline
PFFNet & $\textbf{0.9062 (0.0269)}$ & $\textbf{0.9338 (0.0268)}$ & $\textbf{0.7788(0.0552)}$  \\
\hline
\end{tabular}}
\end{table}

\begin{table}[!htb]
\centering
\caption{The quantitative leaf segmentation results for the CVPPP A1 dataset compared with the state-of-the-art methods.}
\resizebox{0.7\linewidth}{!}{%
\begin{tabular}{l|l}
\hline
Method  & $SBD$ ($\% $ ) \\ \hline
RIS  \cite{romera2016recurrent} & $ 66.6$  \\
RNN \cite{salvador2017recurrent}  & $ 74.7$  \\
Deep coloring \cite{kulikov2018instance}  & $ 80.4$  \\
Embedding-based \cite{payer2018instance}  & $ 84.2$  \\
Discriminative loss \cite{de2017semantic}  & $ 84.2$  \\
Recurrent with attention \cite{ren2017end}  & $84.9$  \\
Data augmentation \cite{kuznichov2019data}  & $88.7$  \\
Harmonic embeddings \cite{kulikov2020instance}  & $89.9$  \\
Synthesis data \cite{ward2018deep}  & $90.0$  \\
\hline
PFFNet  & $ \textbf{91.1}$  \\
\hline
\end{tabular}}
\label{cvppp-challenge}
\end{table}

\begin{figure*}[!t]
\centering
\includegraphics[width=0.72\textwidth]{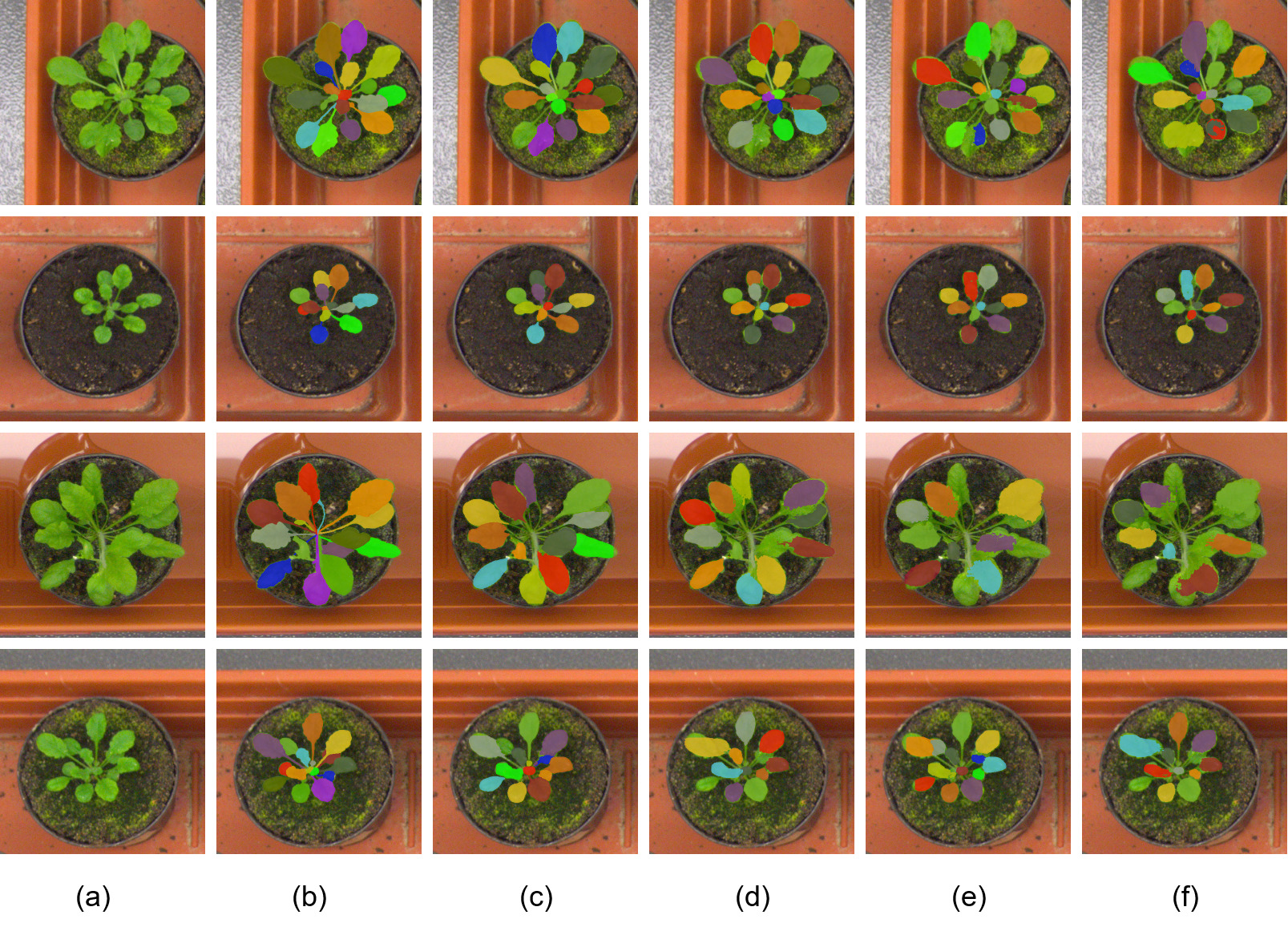}
\caption{The visual comparison of results for CVPPP A1 dataset. (a) original images, (b) ground truth annotations, (c) predictions by our proposed PFFNet, (d) predictions by Cell R-CNN V2 \cite{liu2019nuclei}, (e) predictions by Cell R-CNN \cite{zhang2018panoptic}, and (f) predictions by Mask R-CNN \cite{he2017mask}. }
\label{cmp-vis-cvppp}
\end{figure*}

First, we compare with our previous work, with the 3-fold cross-validation on the $128$ training images. The results are shown in Table \ref{cvppp-cv3-cmp} and Fig. \ref{cmp-vis-cvppp}. By outperforming our previous Cell R-CNN and Cell R-CNN V2 on the instance segmentation tasks for biology images as well as the medical images, our proposed PFFNet is demonstrated to be effective.

To further demonstrate the effectiveness of our PFFNet on biology image analysis, we also conducted a comparison experiment with other previous work, using the $33$ leaf segmentation testing images. Table \ref{cvppp-challenge} is the performance between our work and the state-of-the-art methods and our segmentation accuracy outperforms all the existing published work on this dataset. Among these methods, RIS \cite{romera2016recurrent}, RNN \cite{salvador2017recurrent}, and Recurrent with attention \cite{ren2017end} processed one instance each time, with the help of the temporal chain from recurrent neural work (RNN) or long-short-term memory (LSTM). In addition, \cite{ren2017end} achieved better performance compared with the previous \cite{romera2016recurrent} and \cite{salvador2017recurrent} due to the attention module and proposal-based architecture. As there is actually no temporal information in the leaf instance segmentation task, other methods focusing on the spatial relationship are more suitable for the task with better performance. \cite{kulikov2018instance}, \cite{de2017semantic}, \cite{payer2019segmenting}, and \cite{kulikov2020instance} are proposal-free instance segmentation methods. \cite{kulikov2018instance} divides all the objects into several groups of untouching instances and processes them separately. Instead of directly learning the instance mask prediction for each leaf, \cite{de2017semantic}, \cite{payer2019segmenting}, and \cite{kulikov2020instance} learned high-dimensional embedding maps projected from the original images. Then, clustering algorithms were employed to separate each instance during inference. Without focusing on each object, their performance is still limited due to the lack of local-level information. Similar to our work, the CNN architecture in \cite{kuznichov2019data} and \cite{ward2018deep} are proposal-based Mask R-CNN. With the help of the auxiliary synthesized images, these two methods outperform most of the previous state-of-the-art methods. However, the image synthesis methods in \cite{kuznichov2019data} and \cite{ward2018deep} are entirely based on the characteristic of the leaves in the given plant phenotype images, such as the texture, direction, and spatial relationship with other leaves. Therefore, the methods are task-specific and hard to fit to other datasets with different characteristics. With the help of panoptic-level features in a local and global view, our proposed-based PFFNet outperforms all the other methods on the CVPPP A1 dataset, without any task-specific design. In addition, the competitive performance on other instance segmentation tasks further demonstrates the generalization ability of our PFFNet.

\begin{table*}[!t]
\centering
\caption{The comparison experiments between our proposed PFFNet and the state-of-the-art panoptic segmentation methods for general images.}
\resizebox{0.99\linewidth}{!}{%
\begin{tabular}{c|c|c|c|c|c|c|c|c|c|c|c|c|c|c|c}
\hline
\multirow{2}{*}{} &  \multicolumn{4}{c|}{$Kumar$}  & \multicolumn{4}{c|}{$TNBC$}  &  \multicolumn{4}{c|}{$BBBC039V1$}  &  \multicolumn{3}{c}{$CVPPP$} \\
\cline{2-16}
 & $AJI$ & $Dice$  & $F1$ & $PQ$ & $AJI$ & $Dice$  & $F1$ & $PQ$ & $AJI$ & $Dice$  & $F1$ & $PQ$ & $SBD$ & $Dice$ & $PQ$   \\
   \hline
OGPan~\cite{kirillov2018panoptic}   & $0.5396   $ &  $ 0.7659  $ &  $ 0.6750  $  &  $ 0.4796  $ & $ 0.5350  $ &  $0.7393 $ &  $0.7542 $  &  $ 0.5146 $ &  $ 0.7983  $ &  $ 0.9277  $ &  $ 0.9180 $  &  $ 0.7773$ &  $0.8674   $ &  $ 0.8453 $  &  $ 0.6342  $ \\
  \hline
JSISNet~\cite{de2018panoptic}  & $ 0.5653   $ &  $ 0.7664   $ &  $ 0.7902  $  &  $0.5448  $ & $ 0.5673   $ &  $ 0.7601   $ &  $ 0.7776  $  &  $ 0.5432  $ &  $ 0.8134 $ &  $ 0.9316  $ &  $ 0.9282 $  &  $ 0.7913  $ &  $ 0.8835   $ &  $ 0.8799  $  &  $ 0.6848  $ \\
  \hline
PanFPN~\cite{kirillov2019panoptic}   & $0.5811   $ &  $0.7898 $ &  $ 0.8194  $  &  $ 0.5678  $ & $0.5874   $ &  $0.7715   $ &  $ 0.8129  $  &  $ 0.5720  $ &  $0.8193  $ &  $0.9320  $ &  $0.9275 $  &  $ 0.7960  $ &  $ 0.8871   $ &  $ 0.8905  $  &  $ 0.6999 $ \\
  \hline
AUNet~\cite{li2019attention}   & $0.5898   $ &  $ 0.7904  $ &  $ 0.8056  $  &  $0.5650  $ & $0.5932   $ &  $ 0.7761  $ &  $ 0.8147  $  &  $ 0.5793  $ &  $ 0.8252  $ &  $ 0.9377  $ &  $ 0.9315 $  &  $0.8090  $ &  $ 0.8883   $ &  $ 0.9086 $  &  $ 0.7153 $ \\
  \hline
UPSNet~\cite{xiong2019upsnet}   & $ 0.5797   $ &  $ 0.7904   $ &  $ 0.8315  $  &  $ 0.5667  $ & $ 0.5816   $ &  $0.7703 $ &  $0.8125  $  &  $ 0.5625 $ &  $ 0.8128  $ &  $ 0.9274  $ &  $0.9191 $  &  $ 0.7857  $ &  $0.8902  $ &  $ 0.8941  $  &  $ 0.7180  $ \\
  \hline
OANet~\cite{liu2019end}   & $ 0.5865   $ &  $ 0.7908   $ &  $ 0.8167  $  &  $ 0.5645  $ & $ 0.5933   $ &  $0.7798   $ &  $ 0.8126  $  &  $0.5744  $ &  $0.8198 $ &  $ 0.9372  $ &  $ 0.9330 $  &  $ 0.8085$ &  $ 0.8881  $ &  $ 0.8994  $  &  $ 0.7102 $ \\
  \hline
PFFNet  & $ \textbf{0.6107} $ &  $ \textbf{0.8091}$ &  $ \textbf{0.8345}$ &  $ \textbf{0.5871}$ & $ \textbf{0.6313} $ &  $ \textbf{0.8037}$ &  $ \textbf{0.8600}$ &  $ \textbf{0.6298}$ &  $ \textbf{0.8477}$ &  $ \textbf{0.9478} $ &  $ \textbf{0.9451}$ &  $ \textbf{0.8330}$ &  $ \textbf{0.9062} $ &  $ \textbf{0.9338} $ &  $ \textbf{0.7788} $\\
 \hline
\end{tabular}}
\label{table-cmp-general}
\end{table*}

\subsection{Comparison with State-of-the-art Panoptic Segmentation Methods for General Images}

In Section~\ref{sec-cmp-bio}, we compare our PFFNet with state-of-the-art methods, particularly on biomedical and biological image instance segmentation tasks. To further demonstrate the superiority of our proposed PFFNet, we conduct extensive experiments in comparison with state-of-the-art panoptic segmentation methods, including the original Panoptic Segmentation method (OGPan)~\cite{kirillov2018panoptic}, Panoptic FPN (PanFPN)~\cite{kirillov2019panoptic}, JSIS-Net~\cite{de2018panoptic}, AUNet~\cite{li2019attention}, UPSNet~\cite{xiong2019upsnet}, and OANet~\cite{liu2019end}. Since these methods were originally proposed for general image analysis, we reimplement them on our tasks by following their original paper and source code. For fair comparisons, we maintain the same ResNet101 + FPN backbone and the group normalization strategy for these methods as ours. The inference processes of all the compared methods are also similar to ours, which employ the output of the instance branch as the final predictions. The experiment results are shown in Table~\ref{table-cmp-general}, where our proposed PFFNet outperforms all the compared methods on four instance segmentation tasks for biomedical and biological images.

Among the compared methods, only OGPan~\cite{kirillov2018panoptic} proposes to train the semantic and instance segmentation tasks separately, which achieves the same performance as Mask R-CNN mentioned in Section~\ref{sec-cmp-bio}. By jointly optimizing semantic and instance segmentation branches with a shared backbone, our PFFNet is able to induce the instance branch to process the semantic-level contextual information and therefore achieves better performance than the OGPan method. JSIS-Net~\cite{de2018panoptic} and Panoptic FPN~\cite{kirillov2019panoptic} jointly train an instance branch with a semantic segmentation decoder, without any fusion mechanism between the features from the two branches. By fusing the features for each instance mask prediction to a semantic features map based on its corresponding location prediction, the instance branch in our PFFNet directly involved in the optimization process of the semantic segmentation tasks, which further facilitates the contextual feature learning and achieves better performance.

By integrating the information from the semantic and instance branch, AUNet~\cite{li2019attention}, UPSNet~\cite{xiong2019upsnet}, and OANet~\cite{liu2019end} achieve better performance than Panoptic FPN, JSIS-Net, and OGPan. AUNet~\cite{li2019attention} fuses the features from RPN and instance segmentation predictions with the different stages of the semantic branch via attention mechanism. Although the motivation of the attention feature fusion step in AUNet is similar to our residual attention feature fusion mechanism, the effectiveness of the semantic consistency regularization and the mask quality sub-branch still induce our PFFNet to achieve higher accuracies. For UPSNet~\cite{xiong2019upsnet}, a panoptic segmentation head has been proposed to learn the panoptic-level information from the fused semantic- and instance-level features, which treats each instance as a unique category. However, the scale variation of the objects in the biomedical and biological images makes the size of each instance category vary. This brings the class-imbalance issue for optimizing the panoptic head of the UPSNet and degrades the overall performance. To this end, UPSNet achieves less competitive performance than our PFFNet due to its imperfect design under the biomedical and biological image instance segmentation application settings. In OANet~\cite{liu2019end}, a spatial ranking module is proposed to predict a confidence score for each instance mask prediction during inference, which aims at tackling the objects overlapping issue when small instances stay inside larger ones. However, such a situation might rarely exist in biomedical and biological images, as shown in Fig.~\ref{cmp-vis-tcga},~\ref{cmp-vis-tnbc},~\ref{cmp-vis-fluo}, and~\ref{cmp-vis-cvppp}. Subsequently, the performance gain of the spatial ranking module is limited to our tasks. On the other hand, our mask quality sub-branch is based on the mask segmentation quality for each instance, which avoids the influence from the overlapping situations and maintains the competitive performance of our PFFNet.


\subsection{Ablation Study}

In this section, we conduct ablation experiments on the TNBC, BBBC039V1, and CVPPP dataset, to test the effectiveness of the three newly proposed modules in the PFFNet on different types of images. For TNBC and BBBC039V1, we have the same data settings as the previous experiments. For CVPPP, we conduct 3-fold cross-validation on the $128$ training images.

\subsubsection{Residual attention feature fusion mechanism}

\begin{figure}[!t]
\centering
\includegraphics[width=0.48\textwidth]{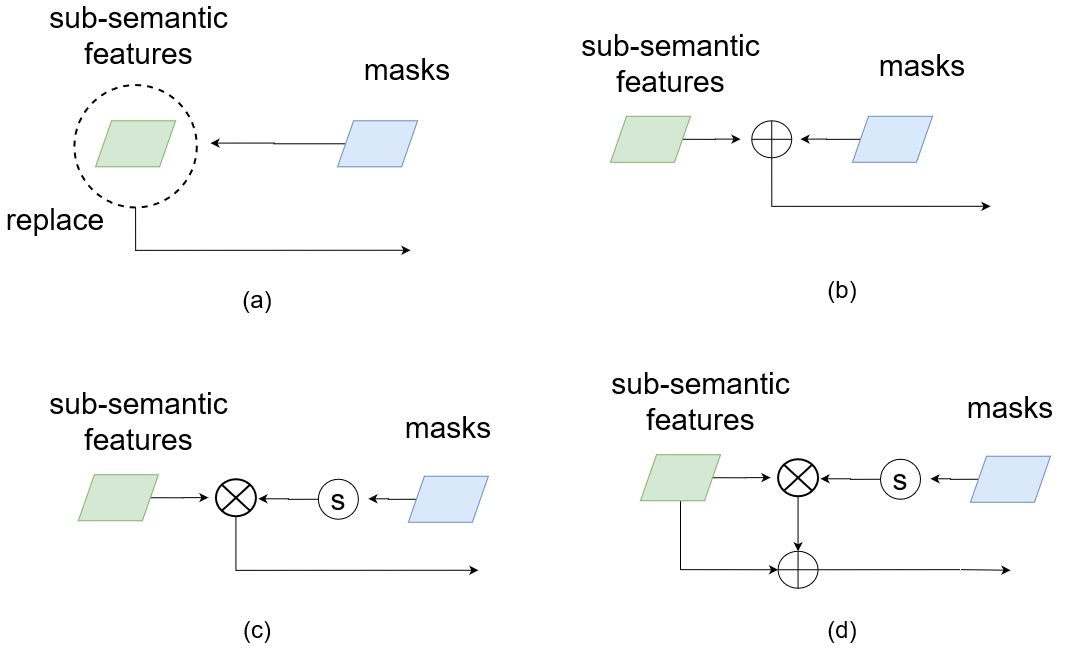}
\caption{Various fusion methods of the proposed residual attention feature fusion mechanism. $S$ represents the sigmoid operation. $+$ is the element-wise summation and $\times$ is the element-wise multiplication. }
\label{abl-atff-fig}
\end{figure}

\begin{table*}[!t]
\centering
\caption{The comparison of results for the different design selections for the feature fusion between the semantic and instance features. The (a), (b), (c), and (d) correspond to Fig. \ref{abl-atff-fig}.}
\resizebox{0.8\linewidth}{!}{%
\begin{tabular}{c|c|c|c|c|c|c|c|c|c|c|c}
\hline
\multirow{2}{*}{} &  \multicolumn{4}{c|}{$TNBC$}  &  \multicolumn{4}{c|}{$BBBC039V1$}  &  \multicolumn{3}{c}{$CVPPP$} \\
\cline{2-12}
 & $AJI$ & $Dice$  & $F1$ & $PQ$ & $AJI$ & $Dice$  & $F1$   & $PQ$ & $SBD$ & $Dice$   & $PQ$  \\
  \hline
(a)  & $ 0.6133 $ &  $ 0.791 $ &  $ 0.8321 $ &  $ 0.5984 $ &  $ 0.8335 $ &  $ 0.9410  $ &  $ 0.9407 $ &  $ 0.8205 $&  $ 0.8947  $ &  $ 0.9200 $  &  $ 0.7491 $\\
 \hline
 (b)  & $ 0.6216  $ &  $ 0.800  $ &  $ 0.8541 $ &  $ 0.6179 $&  $ 0.8420$ &  $ 0.9443 $ &  $ 0.9407 $&  $ 0.8276 $&  $ 0.8983  $ &  $ 0.9274  $  &  $ 0.7603 $\\
 \hline
 (c)  & $ 0.6220  $ &  $ 0.7983  $ &  $ 0.8521 $ &  $ 0.6139 $& $ 0.8403 $ &  $ 0.9443 $ &  $ 0.9427 $&  $ 0.8248 $&  $ 0.8999  $ &  $ 0.9267 $  &  $ 0.7606 $ \\
 \hline
(d)  & $ \textbf{0.6313} $ &  $ \textbf{0.8037}$ &  $ \textbf{0.8600}$ &  $ \textbf{0.6298}$ &  $ \textbf{0.8477}$ &  $ \textbf{0.9478} $ &  $ \textbf{0.9451}$ &  $ \textbf{0.8330}$ &  $ \textbf{0.9062} $ &  $ \textbf{0.9338} $ &  $ \textbf{0.7788} $\\
 \hline
\end{tabular}}
\label{abl-raff}
\end{table*}

\begin{table*}[!t]
\centering
\caption{The ablation study for the mask quality sub-branch. w / o means PFFNet without the mask quality sub-branch.}
\resizebox{0.8\linewidth}{!}{%
\begin{tabular}{c|c|c|c|c|c|c|c|c|c|c|c}
\hline
\multirow{2}{*}{} &  \multicolumn{4}{c|}{$TNBC$}  &  \multicolumn{4}{c|}{$BBBC039V1$}  &  \multicolumn{3}{c}{$CVPPP$} \\
\cline{2-12}
 & $AJI$ & $Dice$  & $F1$ & $PQ$ & $AJI$ & $Dice$  & $F1$ & $PQ$ & $SBD$ & $Dice$ & $PQ$   \\
  \hline
w / o  & $ 0.6150   $ &  $ 0.7960   $ &  $ 0.8495 $&  $ 0.6114 $ &  $ 0.8387 $ &  $ 0.9426 $ &  $ 0.9388 $ &  $ 0.8206 $ &  $ 0.8977   $ &  $ 0.9233  $ &  $ 0.7593 $ \\
  \hline
w  & $ \textbf{0.6313} $ &  $ \textbf{0.8037}$ &  $ \textbf{0.8600}$ &  $ \textbf{0.6298}$ &  $ \textbf{0.8477}$ &  $ \textbf{0.9478} $ &  $ \textbf{0.9451}$ &  $ \textbf{0.8330}$ &  $ \textbf{0.9062} $ &  $ \textbf{0.9338} $ &  $ \textbf{0.7788} $\\
 \hline
\end{tabular}}
\label{abl-qua}
\end{table*}

\begin{table*}[!t]
\centering
\caption{The ablation study for the semantic task consistency regularization. w / o means PFFNet without the semantic consistency regularization.}
\resizebox{0.8\linewidth}{!}{%
\begin{tabular}{c|c|c|c|c|c|c|c|c|c|c|c}
\hline
\multirow{2}{*}{} &  \multicolumn{4}{c|}{$TNBC$}  &  \multicolumn{4}{c|}{$BBBC039V1$}  &  \multicolumn{3}{c}{$CVPPP$} \\
\cline{2-12}
 & $AJI$ & $Dice$  & $F1$ & $PQ$ & $AJI$ & $Dice$  & $F1$ & $PQ$ & $SBD$ & $Dice$ & $PQ$   \\
  \hline
w / o  & $ 0.6206   $ &  $ 0.7974   $ &  $ 0.8370  $  &  $ 0.6043  $ &  $ 0.8390  $ &  $ 0.9430  $ &  $ 0.9416 $  &  $ 0.8244  $ &  $ 0.8995   $ &  $ 0.9267  $  &  $ 0.7601  $ \\
  \hline
w  & $ \textbf{0.6313} $ &  $ \textbf{0.8037}$ &  $ \textbf{0.8600}$ &  $ \textbf{0.6298}$ &  $ \textbf{0.8477}$ &  $ \textbf{0.9478} $ &  $ \textbf{0.9451}$ &  $ \textbf{0.8330}$ &  $ \textbf{0.9062} $ &  $ \textbf{0.9338} $ &  $ \textbf{0.7788} $\\
 \hline
\end{tabular}}
\label{abl-csc}
\end{table*}
 
In this section, we first study the selection of the feature fusion mechanism. As shown in Fig~\ref{abl-atff-fig}, we present $4$ different selections: (a) replacement: replace the semantic feature map with the mask predictions, which is employed in Cell R-CNN V2; (b) summation: sum the semantic feature map with the mask prediction; (c) attention: we first obtain the mask probability maps with the sigmoid operation, then they are multiplied with the corresponding semantic features; (d) residual attention: add a residual connection on (c). For all the experiments, the rest of the model is the same as our proposed PFFNet in addition to the feature fusion mechanism. The comparison results are shown in Table \ref{abl-raff}.

As discussed above, the previous feature fusion mechanism in Cell R-CNN V2 directly replaced the subset of the semantic feature map using the mask predictions for each object, which results in the semantic-level information loss. Therefore, the models with summation (b), attention (c), and residual attention (d) fusion mechanism achieve better performance than the replacement fusion (a), as all of them preserve the original semantic features. Among (b), (c), and (d), residual attention fusion mechanism (d) always achieves the best performance on all datasets, and it is therefore employed in our PFFNet.

\subsubsection{Mask quality sub-branch}

To demonstrate the effectiveness of our proposed mask quality sub-branch, we conducted an ablation study by removing the mask quality sub-branch and comparing the performance with PFFNet. As shown in Table \ref{abl-qua}, the accuracies under all the metrics are decreased after removing the mask quality sub-branch, especially on the object-level metrics. Without the mask quality sub-branch, there exist low-quality mask predictions with a high classification score, which affect the segmentation of the touching object during inferences and are eventually harmful to the object-level performance.

\subsubsection{Semantic task consistency regularization}

Table \ref{abl-csc} illustrates the effectiveness of the semantic consistency regularization by ablating it from the original PFFNet. Although the regularization aims to facilitate the semantic information learning in the instance branch, we notice that the improvements under the object-level metrics are at the same level of the pixel-level metrics, on all the three datasets. Compared with Table \ref{abl-raff} and Table \ref{abl-qua}, we notice that the model without the semantic consistency regularization outperforms the model without the residual attention fusion mechanism and mask quality sub-branch, which indicates the effectiveness of the semantic consistency regularization is the lowest among all the three proposed modules. However, the semantic task consistency regularization is still a novel module as it is implemented by only adding one more loss function, which is straightforward and easy to adapt to other related tasks.

\subsection{Generalization Ability Study}

To demonstrate the generalization ability of our proposed PFFNet, we conduct a generalization study by training the model on one seen dataset and validate it on another unseen one. We follow the experimental setting in~\cite{jahanifar2019nuclick} by training the model on the CPM17 training set~\cite{vu2019methods} and validating it on the TCGA-Kumar testing set. During training, we first randomly cropped $10$ $256 \times 256$ patches from each CPM17 training images. Next, the patches are augmented via horizontal and vertical flipping,  $90^{\circ}$, $180^{\circ}$, and $270^{\circ}$ rotations, Gaussian noise, Gaussian blur, and median blur. For testing, the well-trained model is validated on the $14$ testing images from the TCGA-Kumar testing set. The detailed experimental results are illustrated in Table~\ref{table-cpm17-to-kumar}.

\begin{table}[!htb]
\centering
\caption{The experimental results on the model generalization ability study for our proposed PFFNet.}
\resizebox{0.99\linewidth}{!}{%
\begin{tabular}{l|l|l|l|l|l}
\hline
Method  & $AJI$  & $Dice$ & $SQ$& $DQ$ & $PQ$ \\ \hline
Cell R-CNN V2~\cite{liu2019nuclei}  & $0.5838$  & $0.8006$& $0.7495$& $0.7601$& $0.5713$ \\
\hline
PFFNet  & $0.6048$  & $0.8146$& $0.7559$& $0.7729$& $0.5856$ \\
\hline
NuClick~\cite{jahanifar2019nuclick}  & $0.7940$  & $0.8886$& $0.8001$& $0.9819$& $0.7856$ \\
\hline
\end{tabular}}
\label{table-cpm17-to-kumar}
\end{table}

As shown in Table~\ref{table-cpm17-to-kumar}, our proposed PFFNet achieves better performance than our previous Cell R-CNN V2~\cite{liu2019nuclei} under all metrics. By enhancing the generalization ability of the Cell R-CNN V2, our proposed residual attention feature fusion mechanism, mask quality branch, and semantic task consistency regularization module are further demonstrated to be effective. In addition, we also compare with Nuclick~\cite{jahanifar2019nuclick}, which employs the point annotations of the nuclei instance masks during training and inference. Without accessing the point annotations for the testing images, our PFFNet achieves less competitive performance than Nuclick. However, we notice that the performance of our PFFNet is close to NuClick under Dice score for pixel-level segmentation, and the SQ score for instance-level segmentation. On the other hand, acquiring the nuclei point annotations for the unseen datasets still incurs annotation burdens. If a less competitive accuracy is acceptable, we would prefer not to employ any annotations from the unseen testing datasets, in favour of a low manual annotation cost.

\section{Conclusion}

In this work, we propose a novel Panoptic Feature Fusion Net (PFFNet) for instance segmentation in the biomedical and biological images, which incorporates semantic- and instance-level features. By extending our previous Cell R-CNN \cite{zhang2018panoptic} and Cell R-CNN V2 \cite{liu2019nuclei}, our newly proposed PFFNet is improved with a residual attention feature fusion mechanism, mask quality sub-branch, and semantic consistency regularization. With the help of the residual attention feature fusion mechanism, the semantic features of foreground objects are retained and the mask generators are able to learn more global contextual information from the semantic segmentation task in the instance branch. To alleviate the misalignment issue between the quality of each mask prediction and its classification score, our mask quality sub-branch learns the mask prediction quality scores during training and employs the quality score to re-weigh the classification of each instance prediction during inference. For robust and accurate learning of the semantic features, the semantic consistency mechanism is proposed to regularize the two semantic segmentation tasks jointly. Furthermore, our PFFNet has wide applicability on various biomedical and biological datasets, including histopathology images, fluorescence microscopy images, and plant phenotype images, where we outperform several state-of-the-art methods by a large margin, including our previous Cell R-CNN and Cell R-CNN V2.

By fulfilling the future work in Cell R-CNN V2 \cite{liu2019nuclei}, our PFFNet has been verified to be effective on various biomedical and biological datasets. In future work, we would further adapt our PFFNet to general image processing tasks. As our PFFNet is effective for 2D image analysis, we can also extend it for 3D microscopy image instance segmentation, which is another important and interesting problem related to this work.





{\small

\bibliographystyle{IEEEtran}
\bibliography{IEEEabrv, ref}

}

%








\end{document}